\documentclass[journal,onecolumn,draftclsnofoot,12pt]{IEEEtran}

\usepackage{cite}
\usepackage[pdftex]{graphicx}
\DeclareGraphicsExtensions{.eps,.jpg}
\usepackage[cmex10]{amsmath}
\usepackage{amssymb}
\usepackage{url}
\usepackage{lscape}
\usepackage{longtable}
\usepackage{color}
\usepackage{array,booktabs,arydshln}
\usepackage{lineno}
\usepackage{titlesec}

\usepackage{caption}
\usepackage{subcaption}

\usepackage{svg}

\usepackage{xspace}
\usepackage{eurosym}
\usepackage[capitalize]{cleveref}
\usepackage[binary-units=true,per-mode=symbol]{siunitx}
\DeclareSIUnit\EUR{\text{\euro}}
\DeclareSIUnit{\isotropic}{i}
\DeclareSIUnit{\dBi}{\decibel\isotropic}
\DeclareSIUnit{\ppt}{ppt}

\usepackage[acronym]{glossaries}
\newacronym[plural=AUVs,firstplural=Autonomous Underwater Vehicles (AUVs)]{auv}{AUV}{Autonomous Underwater Vehicle}
\newacronym{mac}{MAC}{Medium Access Control}
\newacronym[plural=ASVs,firstplural=Autonomous Surface Vehicles (ASVs)]{asv}{ASV}{Autonomous Surface Vehicle}
\newacronym{rf}{RF}{Radio Frequency}
\newacronym{csma}{CSMA}{Carrier-Sense Multiple Access}
\newacronym{dacap}{DACAP}{Distance-Aware Collision Avoidance Protocol}
\newacronym{rts}{RTS}{Request-To-Send}
\newacronym{cts}{CTS}{Clear-To-Send}
\newacronym[plural=ACs,firstplural=APOLL Controllers (ACs)]{ac}{AC}{APOLL Controller}
\newacronym[plural=MUs,firstplural=Mobile Units (MUs)]{mu}{MU}{Mobile Unit}
\newacronym[plural=ROVs, firstplural=Remotely Operated Vehicles (ROVs)]{rov}{ROV}{Remotely Operated Vehicle}
\newacronym[plural=USVs, firstplural=Unmanned Surface Vehicles (USVs)]{usv}{USV}{Unmanned Surface Vehicle}
\newacronym{uuv}{UUV}{Unmanned Underwater Vehicle}
\newacronym[]{soa}{SOA}{Service-Oriented Architecture}
\newacronym[]{mavlink}{MAVLink}{Micro  Aerial Vehicle Link}
\newacronym[]{raas}{RaaS}{Robots-as-a-Service}
\newacronym[]{hmi}{HMI}{Human-Machine Interface}
\newacronym[]{ros}{ROS}{Robot Operating System}
\newacronym[plural=UVs,firstplural=Unmanned Vehicles (UVs)]{uv}{UV}{Unmanned Vehicle}
\newacronym[plural=UWSNs,firstplural=Underwater Wireless Sensor Networks (UWSNs)]{uwsn}{UWSN}{Underwater Wireless Sensor Network}
\newacronym[plural=PCBs,firstplural=Printed Circuit Boards (PCBs)]{pcb}{PCB}{Printed Circuit Board}
\newacronym[]{fsk}{FSK}{Frequency Shift Keying}
\newacronym[plural=PDRs,firstplural=Packet Delivery Ratios (PDRs)]{pdr}{PDR}{Packet Delivery Ratio}
\newacronym[plural=PSDs,firstplural= Power Spectral Densities (PSDs)]{psd}{PSD}{Power Spectral Density}
\newacronym{snr}{SNR}{Signal-to-Noise Ratio}
\newacronym{adc}{ADC}{Analog to Digital Converter}
\newacronym{pdd}{PDD}{Packet Delivery Delay}
\newacronym{rmq}{RMQ}{RabbitMQ}
\newacronym{amqp}{AMQP}{Advanced Message Queuing Protocol}
\newacronym{robovaas}{RoboVaaS}{Robotic Vessels as-a-Service}
\newacronym{qoe}{QoE}{Quality of Experience}
\newacronym[plural=SBCs,firstplural= Single-Board Computers (SBCs)]{sbc}{SBC}{Single-Board Computer}
\newacronym[plural=DTNs,firstplural= Delay Tolerant Networks (DTNs)]{dtn}{DTN}{Delay Tolerant Network}
\newacronym{gnss}{GNSS}{Global Navigation and Satellite System}
\newacronym{imu}{IMU}{Inertia Measuring Unit}
\newacronym{csv}{CSV}{Comma Separated Values}
\newacronym{webrtc}{WebRTC}{web real-time communication}
\newacronym{http}{HTTP}{Hypertext Transfer Protocol}
\newacronym{api}{API}{Application Programming Interface}
\newacronym[plural=TRLs, firstplural= Technology Readiness Levels (TRLs)]{trl}{TRL}{Technology Readiness Level}
\newacronym{hil}{HIL}{Hardware in the Loop}
\newacronym{rest}{RESTful}{Representational State Transfer}
\newacronym{ns2}{ns2}{Network Simulator 2}
\newacronym{cidr}{CIDR}{Classless Inter-Domain Routing}
\newacronym{sql}{SQL}{Structured Query Language}
\newacronym{hpa}{HPA}{Hamburg Port Authority}
\newacronym{mbes}{MBES}{Multi Beam Echo Sounder}
\newacronym{wpa2}{WPA2}{Wi-Fi Protected Access II}

\allowdisplaybreaks

\graphicspath{%
     {./figures/}%
}

\newcommand{\fc}[1]{{\color{black} #1}}

\newcommand{\ahoi}{\textsl{ahoi}\xspace}

\renewcommand{\thesubsubsection}{\Alph{subsection}.\arabic{subsubsection}}
\titleformat{\subsubsection}{\normalsize\it}{\thesubsubsection.}{8pt}{}

\newcolumntype{L}[1]{>{\raggedright\let\newline\\\arraybackslash\hspace{0pt}}m{#1}}
\newcolumntype{C}[1]{>{\centering\let\newline\\\arraybackslash\hspace{0pt}}m{#1}}
\newcolumntype{R}[1]{>{\raggedleft\let\newline\\\arraybackslash\hspace{0pt}}m{#1}}
 
\makeatletter
\newcommand\remembertext[2]{
  \immediate\write\@auxout{\unexpanded{\global\long\@namedef{mytext@#1}{\textit{#2}}}}%
  #2%
}
\newcommand\recalltext[1]{%
  \ifcsname mytext@#1\endcsname
    \@nameuse{mytext@#1}%
  \else
    ``??''
  \fi
}
\makeatother

\begin{document}
\title{System Architecture and Communication Infrastructure for the RoboVaaS project - Submitted to IEEE Journal of Oceanic Engineering}
\author{%
    \IEEEauthorblockN{Emanuele Coccolo, Cosmin Delea, Fabian Steinmetz, Roberto Francescon, Alberto Signori, Ching Nok Au, Filippo Campagnaro, Vincent Schneider, Federico Favaro, Johannes Oeffner, Christian Renner, Michele Zorzi}
\thanks{C.~Delea (email: cdelea@cml.fraunhofer.de), Ching Nok~Au (email: ching.nok.au@cml.fraunhofer.de), V.~Schneider (email: vfocke@cml.fraunhofer.de) and J.~Oeffner (email: johannes.oeffner@cml.fraunhofer.de) are with the Fraunhofer Center for Maritime Logistics and Service, Am Schwarzenberg-Campus 4, 21073 Hamburg, Germany.
E.~Coccolo (email: coccoloe@dei.unipd.it), F.~Campagnaro (email: campagn1@dei.unipd.it), F.~Favaro (email: fedefava86@gmail.com), A.~Signori (signoria@dei.unipd.it) and M.~Zorzi (email: zorzi@dei.unipd.it) are with the Department of Information Engineering, University of Padova, 35131 Padova, Italy. 
R.~Francescon (email: roberto.francescon@wirelessandmore.it), is with Wireless and More srl,  35129 Padova, Italy.
F.~Steinmetz (email: fabian.steinmetz@tuhh.de) and C.~Renner (email: christian.renner@tuhh.de) are with the Institute for Autonomous Cyber-Physical Systems, Hamburg University of Technology, Harburger Schloßstraße 28, 21079~Hamburg, Germany.
}
}
\date{}


\maketitle

\begin{abstract}
\renewcommand{\baselinestretch}{1.3}\small
Current advancements in waterborne autonomous systems, together with the development of cloud-based service-oriented architectures and the recent availability of low-cost underwater acoustic modems and long-range above water wireless devices, enabled the development of new applications to support ships and port activities. Unmanned Surface Vehicle (USV) can, for instance, be used to perform bathymetry and environmental data collection tasks to ensure under-keel clearance and to monitor the quality of the water. Similarly, Remotely Operated Vehicles (ROVs) can be deployed to inspect ship hulls and typical port infrastructure elements, such as quay and sheet pilling walls.
In this paper we present the complete system deployed for the small-scale demonstrations of the \gls{robovaas} project, which introduces an on-demand service-based cloud system that dispatches Unmanned Vehicles (UVs) capable of performing the required service either autonomously or piloted. These vessels are able to interact with sensors deployed in the port and with the shore station through an integrated underwater and above water network. The developed system has been validated through sea trials and showcased through an underwater sensor data collection service.  The results of the test presented in this paper provide a proof-of-concept of the system design and indicate its technical feasibility.  It also shows the need for further developments for a  mature technology allowing on-demand robotic maritime assistance services in real operational scenarios.
\end{abstract}

\begin{IEEEkeywords}
\renewcommand{\baselinestretch}{1.3}\small
Unmanned vessels; underwater acoustic networks; Robots-as-a-Service; sea experiments; network performance validation; long range WiFi. 
\end{IEEEkeywords}

\section{Introduction and motivation for this work} \label{sec:intro}

\begin{figure}[t] 
        \centering
       \subfloat[]{\includegraphics[width=0.45\columnwidth]{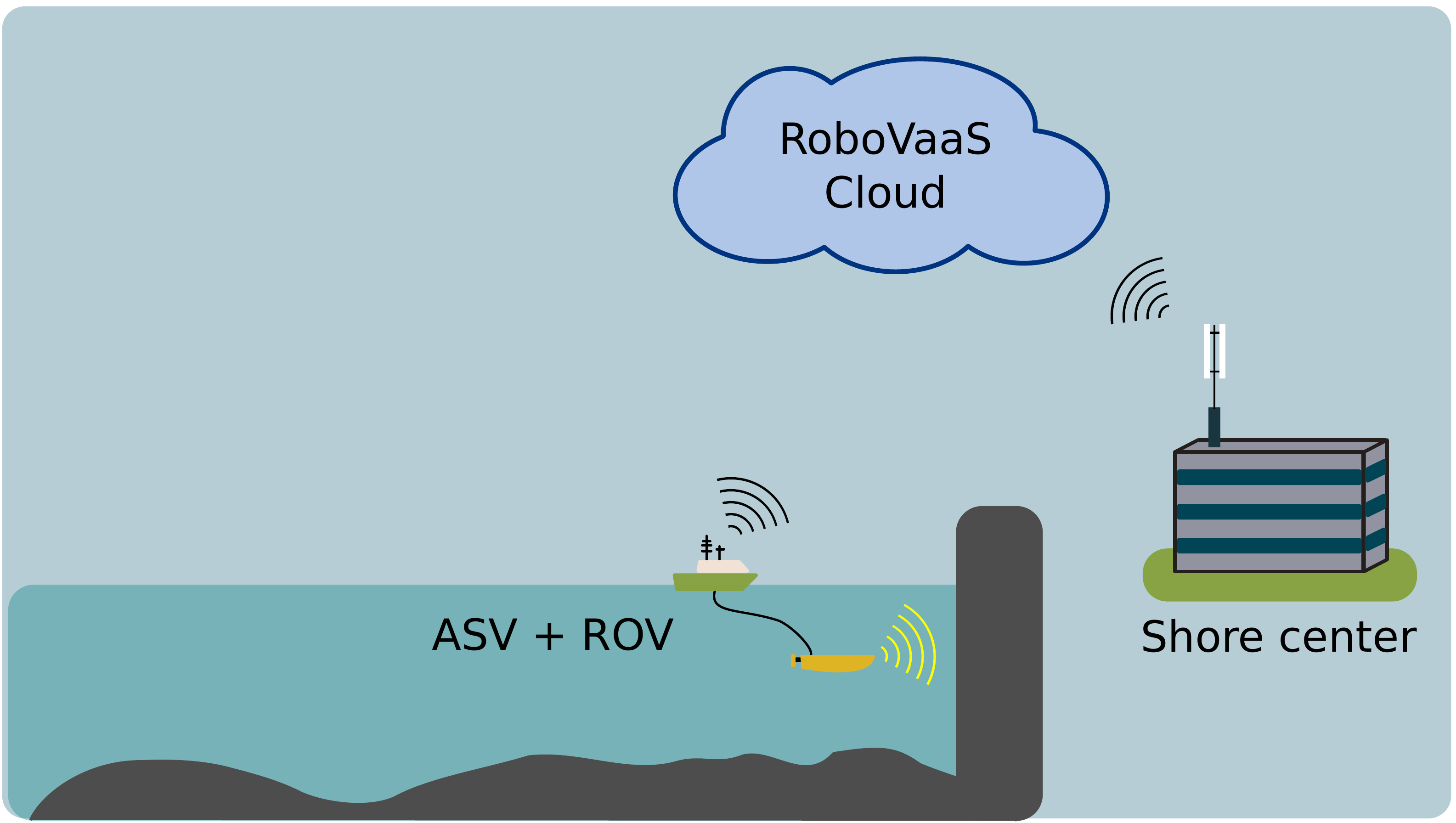}
       \label{fig:quay_wall}}
       \hspace{0.2 cm}
       \subfloat[]{\includegraphics[width=0.45\columnwidth]{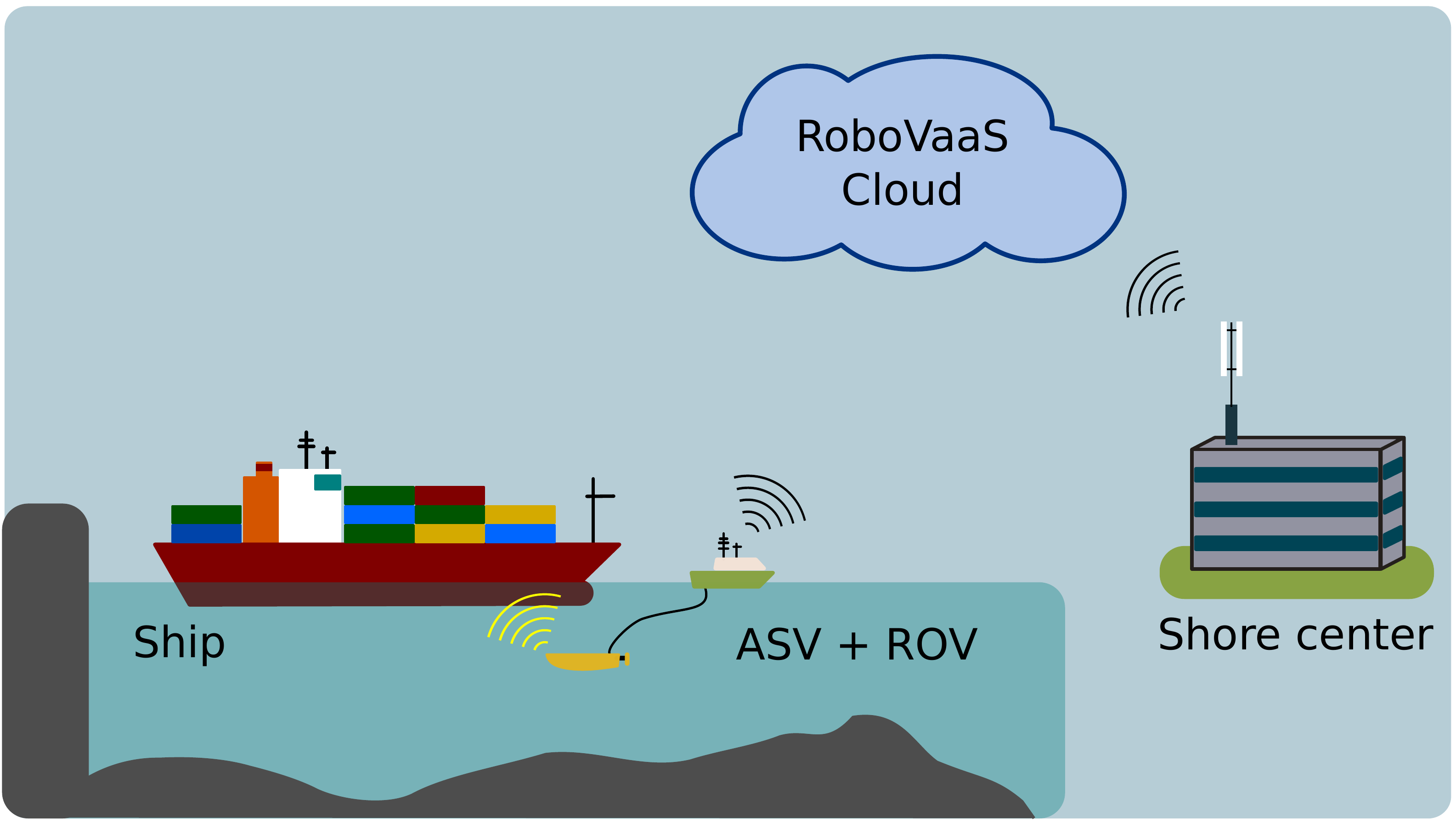}\label{fig:ship_hull}}
       \hspace{0.2 cm}
       \subfloat[]{\includegraphics[width=0.45\columnwidth]{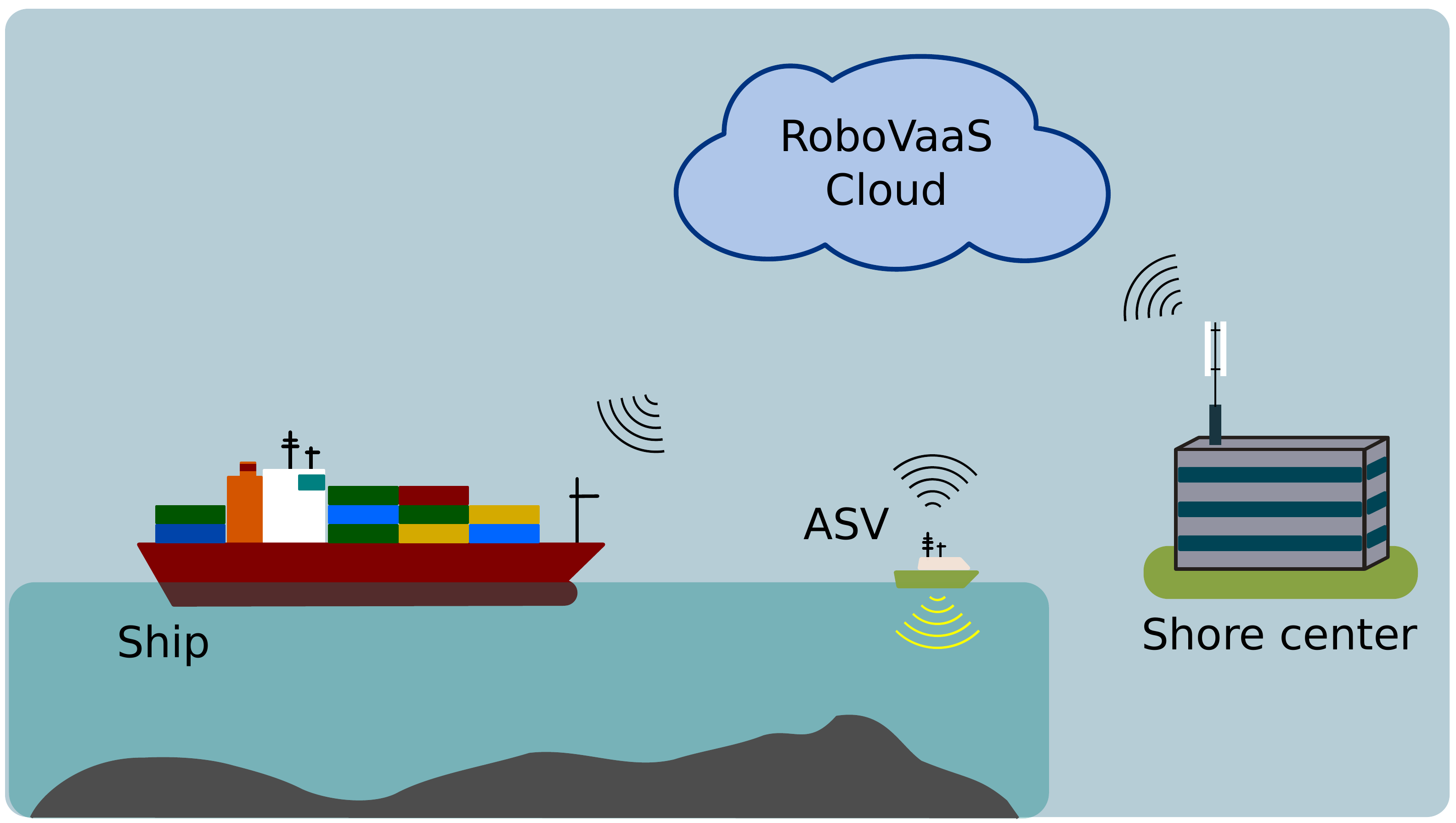}\label{fig:anti_grounding}}
       \hspace{0.2 cm}
        \subfloat[]{\includegraphics[width=0.45\columnwidth]{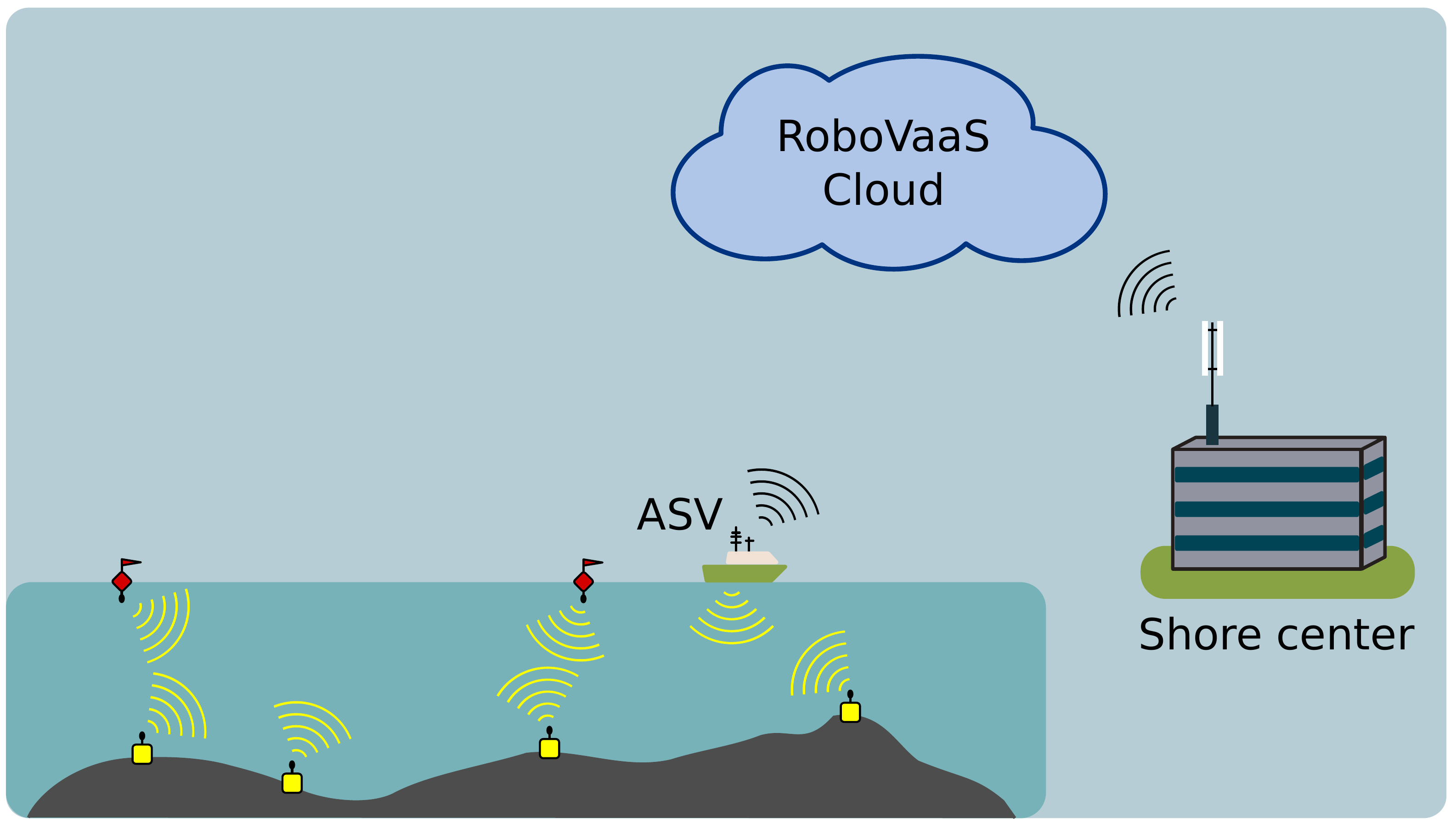}\label{fig:datamuling}}
       \caption{RoboVaaS services: quay walls inspection (a), ship hull inspection (b), anti-grounding (c), environmental and bathymetry data collection (d).}
       \label{fig:robovaas_serveces}
\end{figure}

Applied research in the field of robotics is nowadays trying to find more compelling and user-friendly robotic solutions for enabling various industrial applications in an intelligent manner. The main objectives are increasing efficiency and reducing human hazard. 
In this context, the \gls{robovaas} project~\cite{robovaas} aims to revolutionize the shipping and near-shore operations by offering robotic aided services via interconnected \glspl{uv}, equipped with specialized sensor technology, a reliable data transfer cloud network for above water and underwater communication, a monitoring station, and a real-time web-based user interface.
The high level of autonomy implied in RoboVaaS is expected to be reached by using unmanned vessels such as \glspl{asv} and \glspl{rov}, to perform both autonomous missions and tasks \fc{with a minimal set of commands sent by human operators. For the current work a path follower controller, which follows a two-dimensional vector to a given waypoint, independent of time, is used throughout the in-lab tests and sea trials.} 
The services envisioned in the RoboVaaS project are a ship hull and quay walls inspection service, an anti-grounding service and an environmental and bathymetry data
collection service (Figure~\ref{fig:robovaas_serveces}).

An envisioned user of the \gls{robovaas} system (such as a ship owner, a navigator or a port authority) can use one or more of the aforementioned services by placing specific requests and following their progress through a standard web browser.
Once the service request is issued, the service provider will assign the request in the form of a task or mission to a single or several surface and underwater \glspl{uv}, commanded and supervised from a ground control station. During a mission, its progress and the acquired data are transmitted to the users in real time or - in some cases - after post-processing. Once the mission is completed, the results of the service can be recalled by the user at any point in time, in order to visualize them through a web browser or to generate a report, based on the acquired data.

The popularity of drones, be they airborne, land or waterborne, has exponentially increased in the last years~\cite{drone_popularity}. This is due to the trend of the scientific community and the industry to find consensus for the tools required for operating and developing such robotic systems. The main difficulties consist in the diversity and complexity of the mobile robots, either from the nonlinearities of their mathematical models or from the loosely confined operational envelope. Furthermore, there are great challenges for fusing real-time data and using it as input for complex navigation algorithms or even for teleoperation. The major concern was to minimize the overhead of the exchanged information, which can be done through both hardware and software. For the first option, one can use common field buses that enable high-frequency control operations, i.e., operations that require to transmit more than 200 control messages per second with reasonable latency, \cite{cavalieri1997impact}. The software, instead, can be optimized by choosing more suitable communication protocols for interfacing the robot's subsystems or by using programming languages that are closer to the hardware and give enough flexibility for modeling the architecture. For example, a typical messaging protocol used by drones is \gls{mavlink} which serializes the incoming and outgoing messages into a platform-independent binary format, thus making the communication lightweight and real-time~\cite{koubaa2019micro}. Also popular robotics frameworks such as \gls{ros} make use of UDP-based protocols, such as UDPROS, to decrease latency between peers communicating with the \gls{ros} Master~\cite{igniteacademy}. The drawback is the reliability issues generally related to UDP-based protocols \cite{markovski2001simulation}, especially for datagram streams with high bandwidth needs.
This is particularly important when dealing with high-frequency robots, such as industrial manipulators, that have very strict security requirements and allow a limited number of software interfaces that can be used for interaction with humans. Marine robotics, especially \glspl{asv}, can operate at lower data frequencies as they are slower systems. This adds another degree of flexibility that can be used to cope with the complex system logic expected for commanding such systems for diverse industrial applications.

The unmanned system presented in this paper - the SeaML \gls{asv}¸ - has an operation frequency, defined as the number of JSON-formatted control messages sent every second, of \SIrange[]{1}{10}{\hertz},
that can be transferred over TCP-based protocols, which ensure reliability and lossless data transfer. From the measurements acquired during tests, the size of each JSON control message is less than \SI{1.2}{\kilo\byte}. Usually, such implementations are favored in client-server applications, where the server-side applications are responsible for ensuring a complete and secure data transfer from and to the clients. The information distribution involves packet duplication and routing, which transfers the burden from the unmanned systems to the shore servers, which are easier to scale up. The aim is to maintain the high degree of flexibility required for operating mobile robots but without compromising the security needed for industrial applications. For maritime systems that are aimed to offer industrial-level services, this aspect is crucial, because there are multiple actors (e.g., port authorities, clients, technical operators, etc.) that need to be involved and require a safe and secure infrastructure for on-demand services performed with unmanned systems.

The assumption in this case is that the low operation frequencies and implicit volume of data considered when designing the system architecture do not affect the safety of operation and monitoring of the unmanned system. These assumptions help us build a \gls{soa} which favors the coupling of the robotic system to an industrial ecosystem, aimed at providing on-demand services to the users. This work proposes a marine robotic solution, whose software design is based on \gls{soa}, that enables real-time client-server communication to securely fulfill various on-demand services.

The underwater and above water marine assets used in this project need to exchange information in order to support the \gls{robovaas} services. Acoustic communication systems are used for underwater data transfer, as acoustic transmission and signal processing techniques have reached a remarkable level of maturity~\cite{stojanovic-MC2R-07,Heidemann12,alves16}. New acoustic low-power sensor nodes have been developed~\cite{renner_tosn}, and the data acquired by these sensors has been collected from a newly developed \gls{asv} equipped with both an acoustic modem and an above water wireless link: the \gls{asv} exploits the latter to upload the collected data to shore.

\remembertext{contribution}{\fc{The contribution of this paper is threefold. First, it presents the overall RoboVaaS concept, introducing all robotic-aided services supported by the system. Even though the services provided in RoboVaaS are currently offered by a couple of maritime service providers, there are very few examples of companies providing such a wide spectrum and the technologies chosen for developing them are not publicly accessible. Secondly, the current work describes not only the cloud platform and the user interface employed to require, perform and monitor the services, but also the communication infrastructure and the data flows and interfaces used to convey the data between the unmanned vehicles and the cloud system. As modifications during the deployment phase of a robotic systems can be more expensive than those during the design phase, this work offers a validated  solution that can be further replicated or used as a benchmark. Finally, this work provides the complete details of the deployment of a robotic system performing one of the five envisioned use cases, evaluating the whole communication infrastructure, the functionalities of the principal vehicle and the cloud platform. This paper significantly differs from our previous work~\cite{zordan_robovaas,uwpolling_mdpi,deleacommunication,zugno_robovaas} because previous publications focused on individual aspects of the whole project and evaluated a single component in a simulated or mocked environment, while in this work all components have been integrated together and tested in the field with an actual deployment.}}

\remembertext{novelty}{ \fc{
Two main novel aspects are introduced by this project. The former is the introduction of the ``as a service'' concept to port activities, hence performing those tasks that are usually executed by port authorities periodically, on-demand, i.e., upon request by a registered user that can require the services simply using a web interface. The latter main novel aspect is the fact that all services are performed with unmanned surface and underwater vehicles, hence without the need for manned vessels, specialized crews and/or divers to perform that task, thereby significantly reducing risks for human operators, which increase exponentially with depth. Realizing each service required the implementation of several innovative components, from a surface vessel able to transport an \gls{rov} to the integration of underwater and above water network components and the realization of a very low-cost acoustic modem to retrieve data from a dense underwater sensor network. This paper describes in detail the deployment of a first integrated affordable system that can enable monitoring of coasts, rivers and lakes, at a cost that is an order of magnitude lower than traditional offshore deployments. All components are operational and work in real time, with no need for any further post-processing, proving the high \gls{trl} of our system.
}}

The rest of this paper is organized as follows. Section~\ref{sec:use_case} describes the \gls{robovaas} use-cases and their data streaming requirements, and Section~\ref{sec:relwork} presents the related works of similar projects and solutions to enable the communication of maritime vehicles. Section~\ref{sec:sysview} presents the design of each part of the system, from the architecture of the \gls{robovaas} cloud control system for \glspl{uv}, to the underwater and above water communication infrastructure and the data format used to support the robotics operations. Section~\ref{sec:lab_test} shows the preliminary evaluation of the system in a simulated environment, while Sections~\ref{sec:deploy} and~\ref{sec:result} present the field test deployment settings and its results, respectively. Finally, Section~\ref{sec:concl} draws our concluding remarks.

\section{\gls{robovaas} Use-Cases Description and Requirements} \label{sec:use_case}
Within the high-level \gls{robovaas} vision, a number of services that have a positive impact on near-shore maritime operations have been identified~\cite{deleacommunication}, including a quay walls inspection service (Figure~\ref{fig:quay_wall}), a ship hull inspection service (Figure~\ref{fig:ship_hull}), an anti-grounding service (Figure~\ref{fig:anti_grounding}), and an environmental and bathymetry data collection service (Figure~\ref{fig:datamuling}).
\remembertext{scenarioIdentification}{\fc{ These four services have been identified and defined with the help of the \gls{hpa}, partner of RoboVaaS and crucial stakeholder within the project as a potential end user. In several workshops with an \gls{hpa} liason and HPA departments the consortium was provided the requirements for each service~\cite{robotic_concepts}. For the bathymetry data collection service, a workshop with the hydrography department of \gls{hpa} at an early development stage yielded the necessity of 1) a cleaner webUI Layout, 2) a strict separation of user (job creation) and operator (vehicle control) to avoid uncontrolled vehicle commands, 3) the suggestion of a comment section where users can add job-specific information, and 4) no custom-integration of complex equipment such as \gls{mbes} but use of proprietary software solutions that can be performed on board on separate computing unit and can then relay processed data to the \gls{robovaas} network. Similar feedback loops were performed for the anti-grounding service with nautical officers in state of the art Ship Handling Simulators~\cite{AGS_scenario}.}}
All these services are performed on-demand, upon a request performed by a user registered in the \gls{robovaas} cloud by means of \glspl{uv}, such as \glspl{asv} and \glspl{rov}.
The communication link between the deployed \gls{asv} and the ground control station can be supported either with a dedicated WiFi or WiMAX deployment, or exploiting the cellular coverage provided by an LTE base station located in the proximity of the area~\cite{zordan_robovaas}, if any.

The information flows needed to support all \gls{robovaas} services are:
\begin{enumerate}
    \item a request/response flow between the user, the \gls{robovaas} cloud, and the shore center, which carries the messages for the authentication of the users, the service description, and information about the service availability;
    \item a mission planning flow between the \gls{robovaas} cloud and the \gls{asv}, in which the information about the working area and the details of the mission are sent to the \gls{asv} before the mission starts;
    \item the control and monitoring flow between the unmanned vessels and the control station, which enables the remote control of the system and the execution of the required task;
    \item the after-action report flow, used to upload the outcome of the mission and the data collected during the mission to the \gls{robovaas} cloud.
\end{enumerate}

The control and monitoring flow has different requirements between services, depending on the  service target and layout. In the following we present the details of each service.

\subsection{Quay walls inspection}
For the quay walls inspection service (Figure~\ref{fig:quay_wall}), an \gls{asv}-carried \gls{rov} assesses the status of the quay walls of an area defined by the user.
After the ASV travels to the required area, the \gls{rov} is launched and operated along the quay wall while being supported by the \gls{asv}, which follows its movements.
Both the \gls{asv} and the \gls{rov} are equipped with high-resolution cameras capturing the video for the operator view.

According to both the simulation study and the analysis performed in~\cite{zugno_robovaas}, and to the data stream measurements performed during the project, in order to support the quay walls inspection service, the following transmission streams are
needed:
\begin{enumerate}
    \item two UDP video streams of \SI{1}{\mega\bit\per\second} each to monitor \gls{asv} and \gls{rov} operations with high reliability;
    \item manual command sent over a UDP control stream, with rate \SI{5}{\kilo\bit\per\second}, to control the movements of a small inspection class \gls{rov}~\cite{bluerov};
    \item a manual command sent over a UDP control stream to operate the \gls{asv}, with rate \SI{50}{\kilo\bit\per\second}.
\end{enumerate}

\subsection{Ship hull inspection}

An \gls{asv}-carried \gls{rov} performs the inspection of a ship hull to detect the presence of defects that may compromise passenger safety (Figure~\ref{fig:ship_hull}).
As for the quay walls inspection service, both the \gls{asv} and the \gls{rov} are equipped with high-resolution cameras. Additionally, the \gls{rov} features the Kraken SeaVision system~\cite{seavision}, a laser-based sensor for the 3D imaging of the hull, and multiple sonar cameras.

According to the simulation study and the analysis performed in~\cite{zugno_robovaas}, and according to the measurements performed during the project, in order to support the ship hull inspection service, the following transmission streams are
needed:
\begin{enumerate}
    \item two UDP video streams of \SI{1}{\mega\bit\per\second} each to monitor \gls{asv} and \gls{rov} operations with high reliability;
    \item a UDP control stream to operate the \gls{asv}, with rate \SI{50}{\kilo\bit\per\second};
    \item a UDP control stream, with rate \SI{100}{\kilo\bit\per\second}, to control the movements of a big and stable inspection class \gls{rov} able to operate the 3D laser scanner with high precision~\cite{rov_cris};
    \item a stream for the operational control of Kraken SeaVision 3D mapping system with rate \SI{3}{\mega\bit\per\second};
    \item a stream to upload the 3D images captured by the Kraken SeaVision 3D mapping system. This application produces a large amount of data to be transmitted (that varies depending on the desired resolution): this data can be conveyed to the control station at the end of the mission to avoid overloading the network.
\end{enumerate}

\subsection{Anti-grounding}

The on-demand anti-grounding service envisioned in \gls{robovaas} is meant to deliver real-time bathymetry data to a ship sailing through  narrow riverbeds or shallow waters, allowing it to react to threats not marked in conventional navigational charts or in order to increase sailing intervals in tide dependent waterways. An \gls{asv} traveling ahead of a merchant vessel measures the water depth and thus enables real-time under keel clearance data (Figure~\ref{fig:anti_grounding}) as far ahead as the stopping distance, outperforming commercially available forward-looking sonars in shallow waters. The anti-grounding service is
thus envisioned to guarantee safe navigation in areas where bathymetry data is either unavailable (e.g., in some South American port areas) or outdated (e.g., in the case of the port of Hamburg because of frequent changes in the riverbed due to tidal bores and ship passages which require daily measurements for critical regions~\cite{hpa_water_depth}).

According to the simulation study and the analysis performed in~\cite{zordan_robovaas}, in \remembertext{lora}{\fc{the anti-grounding service the \gls{asv} needs to transmit to the ship a multibeam sonar UDP stream of approximately \SI{0.65}{\mega\bit\per\second}, and can be supported by means of a broadband radio link (either WiFi, WiMax or LTE) plus an additional LoRa backup link used only for alarm messages.}} The \gls{asv} can perform a predefined path with no need for additional data streams. Control and video streams to monitor the mission may be envisioned, but are not strictly required. Within \gls{robovaas}, not only have the technical data link requirements been evaluated, but a nautical simulation using ship handling simulators has also been implemented and nautical officers have executed an exercise to validate the functionality of the prototype~\cite{AGS_scenario}.

\subsection{Environmental and Bathymetry Data Collection}

The environmental and bathymetry data collection service is performed directly by an \gls{asv} moving along a pre-loaded path (Figure~\ref{fig:datamuling}). Both collected data can be either stored in the \gls{asv} and uploaded to the ground control station at the end of the mission, or transmitted to a server in real time, depending on the network coverage.
While the bathymetry data is collected directly by the \gls{asv} through sonar systems, the environmental data can be either collected by the \gls{asv} sensors, or retrieved by the \gls{asv} from submerged sensor nodes collecting measurements for a long period of time. If underwater sensors are deployed in the area where the \gls{asv} is performing any of the aforementioned services, it can collect their data as a side task.
The data is collected from the underwater nodes via acoustic modems, by using a polling protocol. 
From here onward in this paper we refer to the underwater data collection service with the name of data-muling, as the \gls{asv}, acting as a mule, collects the data from the underwater nodes and then conveys it to the shore server. 

The data flow of the data-muling task is related to the sensor data, that needs to be compressed, sent to the \gls{asv} through the underwater network, and finally conveyed to the RoboVaaS server. This data flow does not have stringent requirements in terms of delay or data rate, as each sensor acquires and stores only a few measurements per day (e.g., one sensor measurement every 30 minutes, with a resulting data generation of few kilobits per day), and the data retrieval operation can last until all data is collected from the sensors. 

\remembertext{whyDataMuling}{\fc{Being the latter one of the most challenging scenarios for the communication infrastructure, due to the instability of the underwater acoustic channel and the lack of standard devices for underwater communications, in this paper, after analyzing the \gls{robovaas} system architecture, we present the evaluation of the integrated network in a lake trial where the \gls{asv} performs the data-muling task. Nevertheless, also the feasibility of the other services is assessed with the evaluation of the above water network coverage, throughput, delay and packet delivery ratio.
The evaluations of the \gls{rov}-based use cases have been performed in a separate test and demonstrated with a live demo during the ITS world congress, available online~\cite{its2021}, and is out of the scope of this work.
}}
\section{Related Work} \label{sec:relwork}

During the last decade, several solutions aimed to bring waterborne \glspl{uv} closer to the robustness and security needs of industrial applications have emerged. Both the vehicles themselves and the infrastructure required for secure operation have gone under thorough analysis. For example, in the context of the SUNRISE project~\cite{SUNRISE_presentation_2014}, led by the University of Rome La Sapienza, several testbeds have been built and adapted for static and mobile underwater network testing and experimentation. 
The SUNRISE GATE provides a unified web interface to access every testbed being part of the SUNRISE federation.
The LOON testbed~\cite{loon_2014}, implemented by the NATO STO CMRE, for instance, is envisioned to foster cooperative development of underwater communications and networking  with both mobile and static nodes. This testbed is widely used by the scientific community~\cite{pompili_ofdm,commsnet17,caiti_usbl} to perform tests on underwater acoustic communication and localization. 
The University of Porto, instead, built the UPORTO testbed~\cite{uporto} for collaborative experimentation and synergistic operations with harbor systems, including mapping and ship traffic monitoring and control. This paved the way to the advancement of new projects targeting specific applications; for instance, the works of the CRAS laboratory at INESCTEC, which cover areas such as autonomous navigation~\cite{cras_autonomous_comp}, long-term deployments, data gathering~\cite{barbosa}, mapping~\cite{PINTO202016} and surveillance.

Other laboratories started projects with analogous aims, like the MORUS project, led by the University of Zagreb, which studied the creation of a complex system with aerial and underwater autonomous vehicles~\cite{morus} for security and environmental monitoring.
Similarly, in the SWARMS project~\cite{swarms} several companies and research institutes combined their efforts to make surface and underwater unmanned vehicles cooperate with each other to facilitate offshore operations. 
Another similar project is OceanRINGS~\cite{oceanrings}, where the University of Limerick developed an interface for offshore commercial operations for \glspl{rov}, providing both a real-world and a virtual environment.

Using as-a-service concepts for ports and coastal areas is a new trend offering a certain service on demand without the need to purchase the whole product entities involved. A first draft of the services that then became the core of the \gls{robovaas} project had first been presented in~\cite{jahn2017digitalization} by the Hamburg Port Authority and Fraunhofer CML, in order to obtain a consistent vision of interconnected smart ports. This concept was extended with the four scenarios described in Section~\ref{sec:use_case} and three main development pillars - a small-scale demonstration platform to demonstrate scaled-down versions of the services to the projects stakeholders, the web-based service application design (incorporating the above water and underwater communication) and feedback-based development with stakeholders of potential customers involved~\cite{robotic_concepts}. This work validates the core architecture for performing the aforementioned services that was comprehensively described in~\cite{deleacommunication} and seeks to evaluate the scalability of the solution on a global level.
 
The concept of robotic devices granting services through a centralized system controlled through a web interface was first introduced in~\cite{iot_robot} for a use-case in computer science education. 
In the context of the Ocean Technology Campus Rostock project, the new offshore infrastructure of the Digital Ocean Lab (DOL) will set up an undersea site for testing ideas and simulations under controlled conditions in a real-world environment, where efficient connection of different robotic entities will be demonstrated \cite{ODC}. 
The Robot Web Tools project \cite{toris2015robot} created an open-source framework for communicating with \gls{ros} compliant robotic systems over web sockets, using \textit{rosbridge} servers.

A web-based virtual machine for clients interacting with various \gls{ros} tools has been developed in~\cite{cervera2017cloud} and evaluated during a robotics competition. The successful implementation of this \gls{raas} system has led to the development of \gls{ros} Development Studio \cite{igniteacademy}, that offers users on-demand \gls{ros} tools, such as the Gazebo Simulator, performing the computational workload on the server-side application. Like in \gls{robovaas}, a web interface allows the end-user to access the whole array of services offered by the service provider.
The concept of service robots can also be applied to different domains, such as healthcare. In~\cite{kamei2012cloud}, for instance, the concept of cloud robotics is extended to service robots to assist the elderly and the disabled in their daily activities. The robots are permanently connected to the cloud, thus their operations are monitored by developers that can perform maintenance tasks and provide assistance, when required.

The main differences between the past projects and \gls{robovaas} are the use-cases that have been developed with port-specific operations in mind and a focus on waterborne robotics. 

In the context of data collection from sensor nodes, and more in general in the acoustic underwater environment, the design of the \gls{mac} protocol plays an important role, since packet collisions and subsequent retransmissions may have a strong impact on the network. Indeed, due to the typical low bitrate of the acoustic modems and the high propagation delay, each retransmission significantly increases the channel occupancy. 

In such an environment, in \cite{polling_favaro} the authors compare two \gls{mac} random access protocols against a polling-based \gls{mac} protocol in the data-muling scenario. Specifically, the first random access protocol is a \gls{csma}-based protocol called \gls{csma}-Aloha-Trigger, the second, called \gls{dacap}, is based on \gls{rts} and \gls{cts} signaling. In the \gls{csma}-based protocol the nodes transmit their packets to the \gls{auv} in a \gls{csma}-like fashion only after the vehicle notifies its presence with the transmission of a trigger packet.
In \gls{dacap} sensor nodes are allowed to transmit their data only after the transmission of an \gls{rts} followed by the correct reception of a \gls{cts} packet. The polling-based protocol, instead, is a former version of the protocol used in this paper and presented in \cite{uwpolling_mdpi}. The authors show that the polling-based \gls{mac} protocol in the considered scenario always outperforms the analyzed random access protocols in terms of throughput and packet delivery ratio.

In \cite{apoll} the authors describe APOLL, a polling-based \gls{mac} protocol that groups all the nodes of the network in two classes: the \gls{ac} and the \gls{mu}. In APOLL, the \gls{ac} can decide to schedule a REGISTRATION period in which the \gls{mu} randomly selects a transmission opportunity to transmit its REGISTRATION packet to the \gls{ac}. The number of opportunities and the length of each opportunity in the REGISTRATION period are decided by the \gls{ac}. After the registration, the POLL-REPORT phase starts. In this phase, the \gls{ac} sends a poll to the intended \gls{mu} followed by data for the \gls{mu}, if any, and the \gls{mu} sends a report back to the \gls{ac}. The POLL-REPORT phase goes on until the \gls{ac} decides to schedule another REGISTRATION phase.

\section{System Design and Implementation} \label{sec:sysview}

The \gls{robovaas} system enables users, e.g., port authorities, ship owners, or other port clients, to book different services, monitor their progress and analyze the results through a web-based user interface. This means that as long as the client has access to the Internet, the user can access these services by a standard web browser. In order to enforce the business logic, users who have broker access rights can assign the tasks to operators, which are technicians responsible for deploying the \glspl{uv}. 
\remembertext{BusinessLogic}{\fc{At its core, the \gls{robovaas} system business logic contains a non-\gls{sql} database, with different models developed for each of the four use-cases. Whenever a new mission of any of the developed use-cases is created, it receives a unique identifier, onto which mission critical data is stored. Missions can be created by any type of user, but the real-time connections that allow the flow of data into the database is controlled by users with higher ranking rights, such as operators. The logic enforcement is handled by the controllers of the web applications back-end, which in turn are the single database clients that modify the content of the database, through specific queries. The queries are modeled based on the data inserted by the registered users in the templates handled by the front-end services directly on the web user interface. Content of the template is fed into the controllers of the back-end through specific \gls{api} calls. For each call, the access rights of the user requesting the data is compared to the access right table in the database. The latter can be modified in the same manner, but solely by the users with administrative rights. While the structure itself is not entirely a novelty, the key elements that differentiate the \gls{robovaas} business logic have been explained in \cite{deleacommunication} and mainly refer to its applicability to waterborne robotics deployed in industrial sites, where little to no drone-specific infrastructure is provided. The presented solution can be first of all deployed in a mobile version, where all hardware is set on-shore, next to the place where the port authorities are performing their daily duties. Moreover, through usage of broadband network connectivity, the solution can be up-scaled with dedicated data centers, where a data warehouse system can gather and analyze data from various \gls{robovaas} units deployed on site. 
Currently, the \gls{robovaas} service architecture is not integrated into the operational structure of any port, as it is under development. But it can be imagined that, once the system is approved, it can be easily integrated into the authorities control environment by accessing the webUI as an administrator. This way, possible users inside the port administration, e.g., the bathymetry department, can be assigned permanent roles while temporary users such as owners of visiting ships can request a time restricted access to book an anti-grounding service or an inspection service.}}
The operators are also responsible for choosing the \gls{uv} available to perform the service and ensuring that the legal and safety procedures for performing such operations are met. A communication channel between the \gls{uv} and the control infrastructure is established only if the following procedures are carried out: an operator has been assigned to perform the task, a free \gls{uv} ready to be deployed is selected by the assigned operator, and the operator has confirmed starting the task.

\subsection{System Architecture}\label{sec:sys_architecture}

\begin{figure}
    \centering
    \includegraphics[width=0.9\columnwidth]{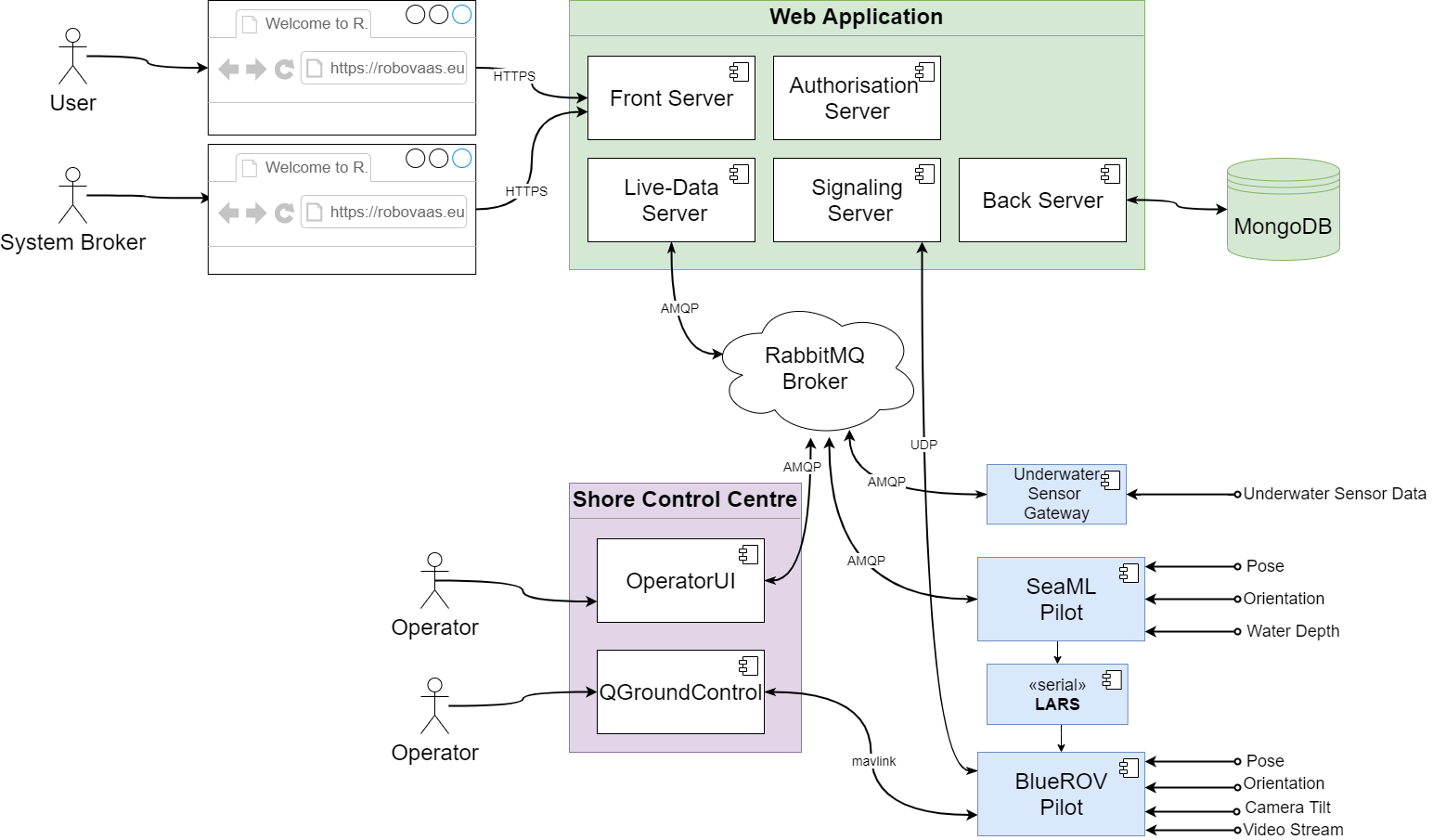}
    \caption{\gls{robovaas} System Architecture: Users and System Brokers use the web interface to connect to the Front Server, which in turns redirects all requests, made through interaction with the user interface, to the responsible server. Operators use dedicated software for direct command of the robotic system. The live data is exchanged through \gls{rmq}, which regulates the access rights to the queues. The continuously changing access rights to the live data are managed by a dedicated server, that oversees the ongoing missions and actors.}
    \label{fig:system_architecture}
\end{figure}

The overall system, depicted in Figure~\ref{fig:system_architecture}, is decomposed into multiple client applications and server-side microservices. The core subsystems that are used for performing any of the proposed on-demand services are the following:
\begin{itemize}
    \item the web application: a collection of Node.js web servers that authorize users to insert or retrieve relevant information into or from a database. Moreover, they implement asynchronous messaging protocols such as SocketIO or \gls{amqp} not only for enabling the client-server API, but also for consuming live data coming from the \glspl{uv};
    \item \gls{rmq} Broker for efficient distribution of real-time data between the robotic system and the human operator or web-based clients;
    \item a \gls{webrtc}-based server for real-time delivery of video packets from \glspl{uv} to web-browser clients;
    \item a file server for reviewing high-quality video footage;
    \item standalone applications, distributed along some \glspl{sbc} and comprising the control system of the \glspl{uv} and their clients having direct access to their operators' inputs.
\end{itemize}

\subsection{System Functionality}

The \gls{robovaas} system is intended to be commanded within a local area network, through operator applications offering a remote \gls{hmi} to the deployed robotic system. Remote users can interact with the \gls{robovaas} system by generating requests to handle specific missions on a web-based user interface. The latter requires nothing else than a standard internet browser to use the application. Provided that their user account exists and has the permissions to generate requests for one or more use-cases, they will automatically create an entry in the job lists of the central database. 
Each user will register and authenticate through a standalone application or through the browser-based client, which will send a request to the Back Server, which has exclusive reading and writing access to the central database. The request is executed by using a \gls{rest} \gls{api}. Once authenticated, the user receives a unique user identification number (userID) and can generate any number of task requests for the service types that have been granted. In the \gls{robovaas} vision, specific service types can be granted to users based on the type of subscription they possess. Information such as task descriptions, user roles and the acquired data is safely logged into a MongoDB, which is a non-SQL database. The Back Server enforces business logic and delegates the job of verifying authentication tokens to the authentication server. MongoDB was preferred to more standard SQL databases due to its flexibility for storing data and hence shorter development time. All data is encoded in JSON format, which allowed a loose definition of the table structure and easy modification of the expected input based on the use-case, without redesigning the relational diagrams or refactoring the database structure. The drawback is the limitation in terms of querying data belonging to various entities (e.g., multiple robots) and lumping the information efficiently. This could become a problem if more complex relationships between tables were expected, in which case relational databases would be preferable.
Additionally, live data is handled through an \gls{rmq} Broker, which binds the queues of the parties for packet multiplication and routing. While some user types, such as the \gls{asv} or the operator, have for safety reasons permanent access rights to the \gls{rmq} exchanges and queues, others, such as web-based clients, have their account dynamically generated by the Back Server and only consume live data. The business logic forces each client to generate its queues following a specific format and to bind them to the exchanges only when the task is started by the operator and confirmed by the Back Server. 
Should a third party try to listen to the exchange handling the communication between the operator and the robot, its access would be immediately denied by the \gls{rmq} Broker, as it had not received permission for its user from the Back Server. In that sense, the Back Server makes use of the \gls{rmq}'s \gls{http} \gls{api} to set the access rights accordingly.

\begin{figure}[ht]
    \centering
    \includegraphics[width=0.8\columnwidth]{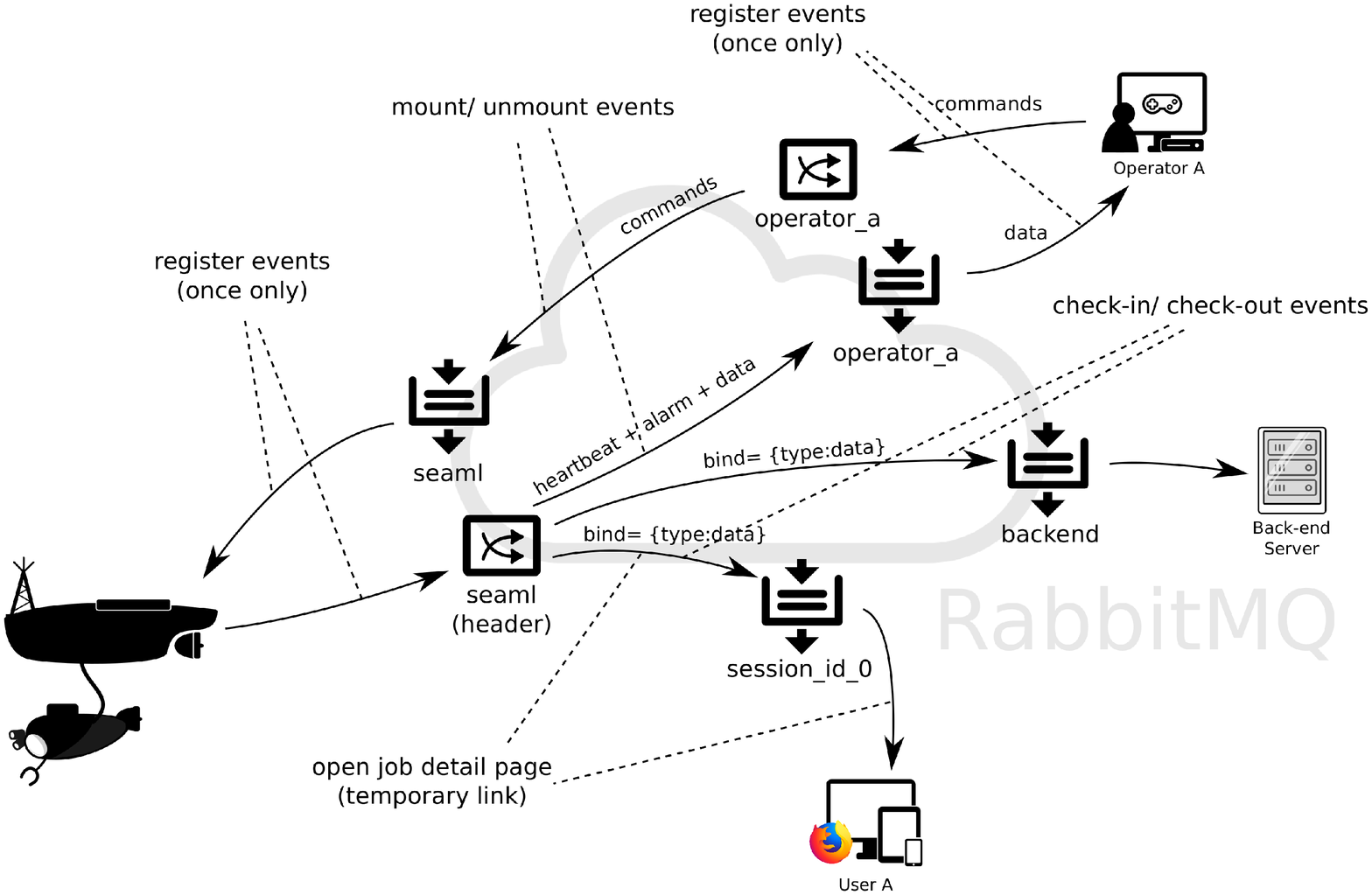}
    \caption{Access Control Scheme for \gls{robovaas} Application: users interacting from the web user interface are allocated temporary queues by the \gls{rmq} server for listening to live data. In parallel, queues are generated for the operators, whom are also given access keys to write into the robotic system queues, where commands are submitted.}
    \label{fig:access_control}
\end{figure}

The summary of the access rights control is depicted in Figure~\ref{fig:access_control}.

\subsection{Underwater Network}

An underwater network enables the communication between \glspl{asv} and battery-powered underwater sensors, used to collect environmental data to monitor a certain area.
Indeed, electromagnetic signals propagate only up to a few meters under water, while optical communications are affected by sunlight noise and water turbidity, providing a very short range in typical shallow water scenarios~\cite{oceans2015genova_multimodal}. Acoustic signals, instead, can propagate up to several kilometers under the sea, at the price of a low data rate, large propagation delay, and poor performance in case of multipath, high shipping activities, or noise caused by wind waves~\cite{stojanovic-MC2R-07} or snapping shrimps~\cite{mandar_shrimps}. In addition, acoustic communication with mobile nodes is strongly affected by the Doppler effect~\cite{renner_tosn}.

In this project, the underwater communication is set up with the \ahoi low-cost and low-power acoustic modems~\cite{renner_tosn}, and a complete communication stack implemented using the DESERT Underwater Framework~\cite{ucomms16_desert}. 
DESERT is a network simulation and emulation tool, based on NS-MIRACLE~\cite{miracle-eurasip}, which implements a complete ISO/OSI modular stack for underwater communications on top of \gls{ns2}. An important feature of this framework is the \textit{RealTime scheduler}, which overwrites the existing \gls{ns2} scheduler and synchronizes the program with the machine internal processor, so that the simulator can operate in real time with other programs or external devices while using the same protocols implemented for simulations.

Concerning the transmission of packets through the medium, a physical module was implemented to act as driver with the \ahoi~modems: more details on the implementation can be found in~\cite{ucomms21_desert}.

\subsubsection{Uwpolling Protocol} \label{sec:uwpolling}

\begin{figure}[t] 
        \centering
       \subfloat[]{\includegraphics[width=0.6\columnwidth]{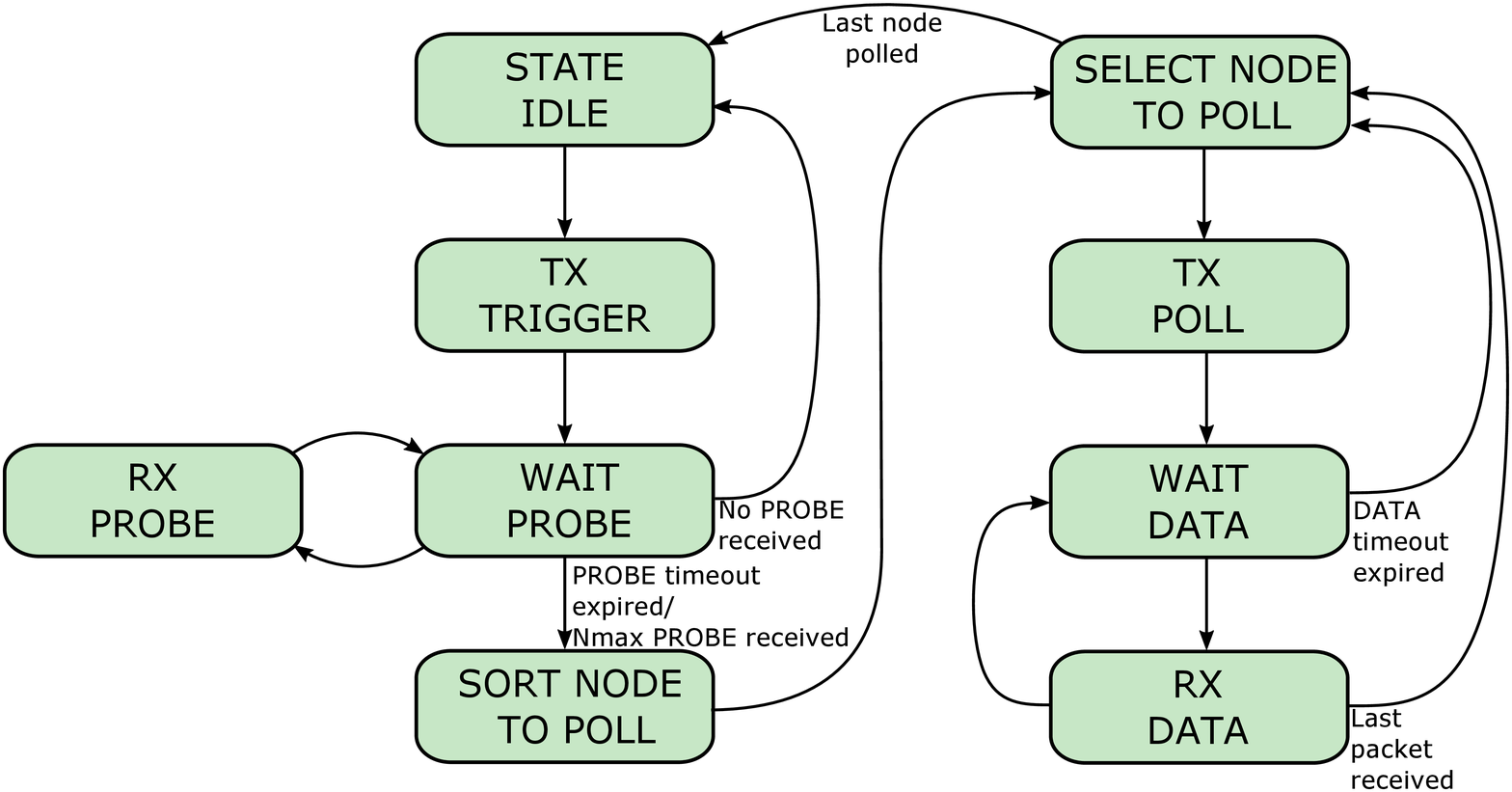}
       \label{fig:AUV_polling}}
       \hspace{1.5 cm}
       \subfloat[]{\includegraphics[width=0.225\columnwidth]{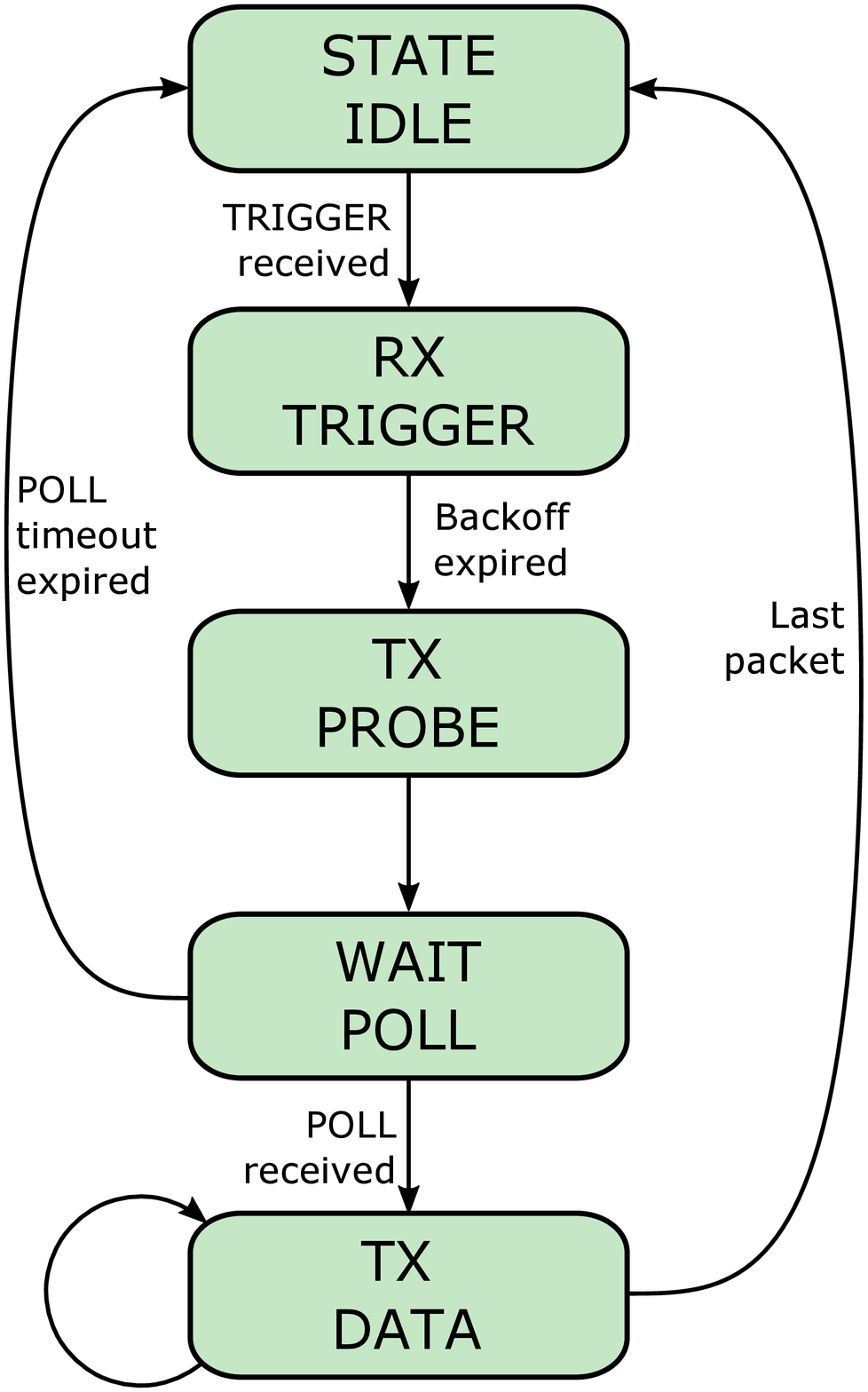}\label{fig:NODE_polling}}
       \hspace{0.4 cm}
       \caption{State machine of the vehicle (a) and sensor node (b) for the \textit{UWPOLLING} \gls{mac} protocol implemented in the DESERT Underwater framework.}
       \label{fig:uwpolling_state_machine}
\end{figure}

As previously stated, the underwater network is used to collect data from sensor nodes. Specifically, during the underwater data collection service, an \gls{asv} patrols the selected area to gather the data from the underwater sensors deployed in the network. For this purpose, \textit{UWPOLLING}~\cite{uwpolling_mdpi}, a polling-based \gls{mac} protocol, has been designed to guarantee fair channel access to all nodes. The protocol assumes up to 3 types of nodes in the network: 
\begin{itemize}
    \item the sensor nodes, which are equipped with sensors to collect data from the surrounding environment;
    \item the vehicle acting as a mule, that can be either an \gls{asv} or an \gls{auv}, which is in charge of moving around in the network and collecting the data from the sensors;
    \item the sink node, which collects data from the vehicle and forwards it to the shore through an above water \gls{rf} link.
\end{itemize} 
Depending on the scenario, the vehicle can also act as the sink of the underwater network, e.g., an \gls{asv} equipped with both acoustic and \gls{rf} modems capable of collecting data from underwater sensors and directly forwarding it to the shore. 
Being this the configuration used in the lake test, in Figure~\ref{fig:uwpolling_state_machine} we present the state machine of the underwater sensor node and of the mobile node, i.e., the \gls{asv}. The complete protocol state machine when the system is used with an external sink is presented in \cite{uwpolling_mdpi}.
The polling protocol works in two subsequent phases: the discovery phase and the polling phase. The first is used by the \gls{asv} to let the other nodes know about its presence, the second is employed for the data collection from the sensors and eventually to forward the data to the sink.

The discovery phase starts when the vehicle sends a TRIGGER packet to the surrounding nodes. The sensor nodes and the sink that correctly received the packet from the \gls{asv} can decide whether to reply with a PROBE packet, based on their availability of data to transmit. If the node has data to send to the vehicle, it transmits the PROBE after a random backoff uniformly chosen between a minimum ($T_{b_{min}}$) and a maximum value ($T_{b_{max}}$). The upper and lower limit of the backoff value are inserted in the TRIGGER packet and can optionally be adapted by the \gls{asv} based on the estimated node density \cite{uwpolling_mdpi}. To let the \gls{asv} know the amount of time to allocate for each of the nodes, the number of packets the node is going to transmit in the polling phase is inserted in the PROBE packet. 

After the reception of the PROBE packets from the surrounding nodes, the \gls{asv} starts the polling phase by creating the poll list. The list provides the order in which the nodes will be polled by the \gls{asv}, i.e., the order in which each node will transmit its data. The poll list is created based on proportional fair scheduling, that aims to provide almost the same possibility of transmission to all nodes, but without penalizing too much the overall throughput of the network \cite{FairScheduling}. Indeed, proportional fair scheduling decides the node order based on both the packets transmitted so far by each node 
and the number of packets a node is going to transmit 
(more details can be found in \cite{uwpolling_mdpi}).
Once the poll list is created, the vehicle starts to poll one by one all the nodes in the list. Specifically, the \gls{asv} sends a POLL packet to the first node in the list, and waits for data from that node. The intended node, once it has received the POLL, sends the data packets to the \gls{asv}. The vehicle waits until either all the intended packets have been received or a timeout related to the amount of time allocated for that node expires. 
Then, the \gls{asv} starts to poll the next node in the list up to the last one. 
Once all the nodes have been polled the discovery phase starts again.

\subsubsection{\ahoi Modem}

To perform the underwater acoustic transmissions, we selected the \ahoi modem, a low-cost and low-power acoustic underwater modem~\cite{renner_tosn}. The modem was developed for the integration into micro \glspl{auv} and \glspl{uwsn}. It consists of three stacked \glspl{pcb} with a size of 50x50x25~mm and an external commercial hydrophone. For the underwater nodes the Aquarian Audio~AS-1 hydrophone~\cite{Aquarian:Hydrophones} was used, which is the default hydrophone for the \ahoi modem. The price of a single \ahoi modem is around~\SI{600}[\EUR]{} (\SI{200}[\EUR]{} for \gls{pcb} and component costs and~\SI{400}[\EUR]{} for the hydrophone). 
Furthermore, the \ahoi modem uses a \gls{fsk} based data transmission with a~\SI{25}{\kilo\hertz} bandwidth centered at~\SI{62.5}{\kilo\hertz}. The transmission range between \SI{50}{\kilo\hertz} and \SI{75}{\kilo\hertz} was selected based on the hydrophone characteristics. In addition, the communication band is above the frequencies that contain most of the acoustic noise produced by near vessels and \glspl{auv}~\cite{SMP_OCEANS2019_ModemResilience}.

\subsection{Above Water Network}\label{sec::wifinetwork}

\remembertext{mikrotikdesc}{\fc{The above water section of the network has been designed and implemented using equipment manufactured by Mikrotik~\cite{mikrotik}.
Hereafter, we give a brief description of the topology of the above water network deployed.
On the \gls{asv}, a Mikrotik Metal 52AC WiFi CPE~\cite{metalac} has been adopted. The Metal 52AC device is a router, mounting a Gigabit Ethernet port, a WiFi 802.11b/g/n and 802.11ac compliant WiFi radio (configured to act as a client) and a \SI{6}{\decibel\isotropic} omnidirectional antenna. The Metal device has been connected with an Ethernet cable to the \gls{asv}'s hardware used to forward the data acquired from the underwater network to the above water network.
On the pier, we used mANTBox 2 12s~\cite{mantbox} as the WiFi AP, which mounts a \SI{12}{\decibel\isotropic} \SI{120}{\degree} directional antenna and an 802.11b/g/n WiFi modem, able to cover long distances and challenging radio channels. We employed a 5 Ethernet ports Mikrotik hEX~\cite{hex} as the core router, which connects the mANTBox, the RabbitMQ servers and, optionally, laptops for troubleshooting and monitoring, thus allowing the end-to-end connectivity between the Tinkerboard and the servers.
We preferred to design a flat network, with Mikrotik hEX acting as the core router and providing  a DHCP server for the entire network, assigning a /24 \gls{cidr} for the entire network. The entire network is, then, on the same ``layer 2 domain'' avoiding the burden of static routing configuration on hEX. The assigned subnet is large enough (254 addresses available) to be scalable in this scenario and to allow many laptops to connect for monitoring and troubleshooting, using both the WiFi network created by mANTBox (in-band with data transmission from Metal AC) and dedicated Ethernet ports on hEX. In order to extend the number of available Ethernet ports in this case, an un-managed switch connected with Mikrotik hEX can be used.}}

\subsection{Data Compression and Live Data Generation}\label{ch:data_compr}

In addition to DESERT, other two programs have been implemented for the underwater telecommunication pipeline, namely:
\begin{itemize}
    \item DATA\_SENS, an application that acquires data from either a real or a mocked sensor, formats it to a compressed string, and sends it through the DESERT application layer;
    \item NET\_BRIDGE, the application used in the \gls{asv} to forward the collected sensor data from the underwater network to the \gls{robovaas} Cloud.
\end{itemize}


DATA\_SENS interfaces with the DESERT Framework at the application level through a module that generates a socket connection, aiming to communicate with an external program that generates the data to be transmitted following the protocols in DESERT.

The main reason DATA\_SENS sends the data through the acoustic channel with a formatted and compressed string with a fixed size rather than standard JSON formatted strings, is due to the limitation of the acoustic channel, that allows only the transmission of small data packets.


DATA\_SENS, therefore, creates a string for each sensor measurement formatted as presented in Table~\ref{tab:sensor_string}. The first 2~bytes of the string compose the ID of the sensor node that generated the data, the following 6~bytes the timestamp at which the data was generated, formatted in hours minutes seconds, one byte is used to represent the datatype (temperature, salinity, etc.) acquired by the sensor, and, finally, the remaining bytes are used to represent the data value. For instance, in the case of the temperature data type, 4 bytes are used for the value: with this data compression, the underwater nodes transmit temperature data with a packet of 25 bytes, including the header introduced by DESERT. For other data types, the length depends on the number of bytes used to represent the value.

\begin{table}[]
    \centering
    \caption{DATA\_SENS string format} \label{tab:sensor_string}
    \begin{minipage}{\linewidth} \centering
        \begin{tabular}{ccccc}
 \textbf{Parameter} & ID      & TIMESTAMP &  DATATYPE & VALUE \\
 \textbf{Format} & \texttt{XX}& \texttt{HHmmss}    &   \texttt{\textless{}type initial\textgreater{}} &    \texttt{\textless{}value\textgreater{}} \\
 \textbf{Size [bytes]} & 2 & 6 & 1 & 4\footnote{4 bytes are used for a temperature sensor measurement, other sizes might be used for other data types.} 
        \end{tabular}
        \end{minipage}
\end{table}

NET\_BRIDGE, instead, parses the sensor packets received by the \gls{asv}, and converts them into a JSON file, that is then sent to the \gls{robovaas} cloud by using an \gls{amqp} client that transmits the JSON file by using the above water network.

For example, the string sent by node 16 through the underwater network at time 14:23:45, with temperature data with value \SI{18.5}{\celsius}, would be \texttt{16142345T18.5}, which is translated at the \gls{asv} into this JSON file, 
where the date is either the current date or the day before, as we assume that the data has been generated no more than 24 hours before the current time:

\texttt{ \{ }

\texttt{\qquad"buoy\_id" : "16", }
 
\texttt{\qquad"data\_type" : "temperature", }
 
\texttt{\qquad"recorded\_at" : "2021-03-02T14:23:45Z", }
 
\texttt{\qquad"value" : 18.5 }

\texttt{ \} }

\section{Preliminary In-Lab Trials}
\label{sec:lab_test}
In order to perform some preliminary tests of the system and evaluate its functionality before the actual sea trial, we performed some in-lab \gls{hil} simulations using a digital twin of the \gls{asv}.
This methodology allowed us not only to test each component of the system separately, but also to perform an integration test of all software and hardware components not directly related to the vehicle's control and navigation system, with consequent reduction of testing costs and time, thereby significantly reducing the probability of failure during the final sea trial demonstration.
The model-based designed techniques make use of simulators for low-cost design validation. In addition, communication between different subsystems is abstracted within the simulator environment, allowing for realistic testing and optimization of the system. Consequently, we developed a digital twin of the real system for both the carrier \gls{asv} and the underwater equipment.

\begin{figure}[htbp] 
       \centering
       \subfloat[SeaML \gls{asv}]{\includegraphics[width=0.49\columnwidth]{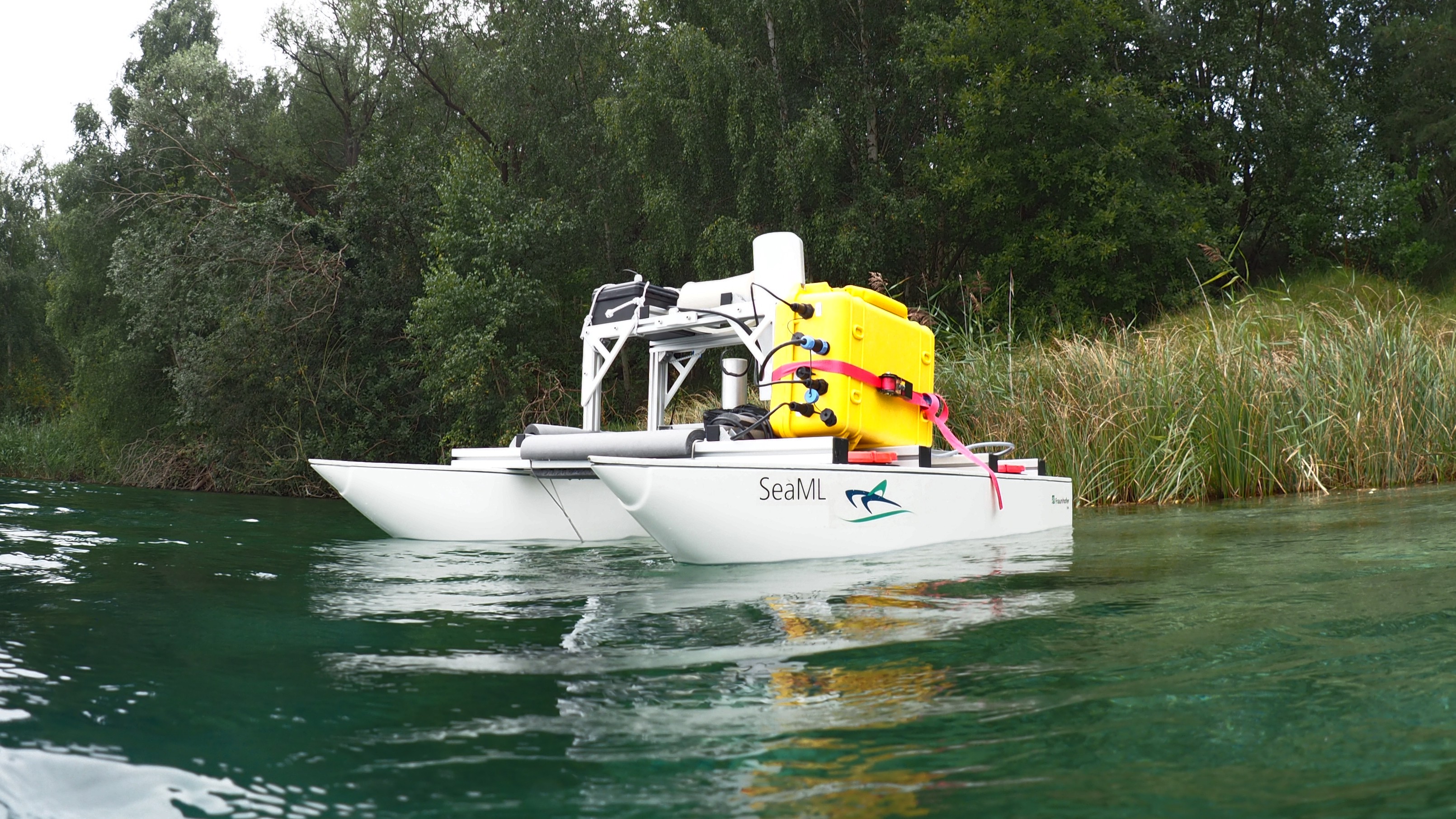}}
       \hspace{0.1 cm}
       \subfloat[Heron digital twin]{\includegraphics[width=0.49\columnwidth]{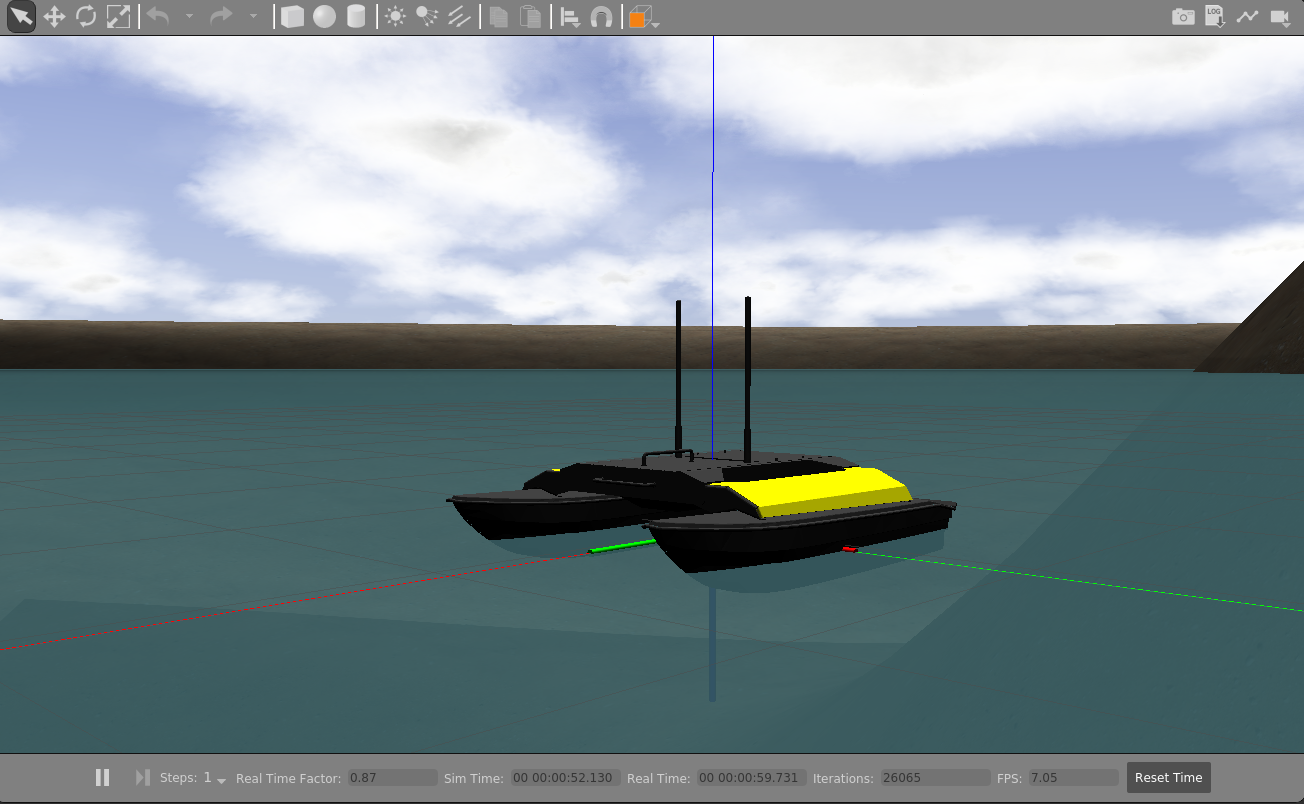}}
       \caption{\gls{asv} simulation setup: thanks to the very similar physical properties, the Heron \gls{asv} has been used for in-lab tests. By abstracting the hardware of the ship, most of the RoboVaaS system components could be cost-effectively tested. Underwater communication is separately simulated and connected to the RoboVaaS system via publicly accessible \gls{rmq} server.}
       \label{fig:simulation_gazebo}
\end{figure}

The Gazebo~\cite{gazebo} simulator was used to simulate the full-body dynamics of the \gls{asv} employed to perform the data-muling exercise, while DESERT was used to emulate the exact behavior of the underwater network that acts as in a real exercise. For the \gls{asv} model we opted for an open-source Heron~\cite{heron} 
model from ClearPath Robotics. The model, depicted in Figure~\ref{fig:simulation_gazebo}, 
makes use of Gazebo's UUVSim~\cite{uuvsim} 
plug-in for rendering the hydrodynamic forces exerted upon the \gls{asv}. Assuming that the physical properties of Heron are reasonably close to the SeaML \gls{asv}, the Heron model was used without any adaptation. The development of a digital twin of the SeaML \gls{asv} is out of scope and will be sought in further works, by changing the model definition 
in the \gls{asv} description package. The trade-off is the need to tune the weights of the course controller to the Heron \gls{asv} while not having any guarantee of similar performance for the SeaML \gls{asv}. The controller output is redirected to the motor set points of the Heron \gls{asv} and the outputs of the emulated sensors, specifically the \gls{gnss} positions and filtered \gls{imu} values, are read by the position and course controller, respectively. 

Because the low-level controller is platform-independent, the companion software as well as the operator application could be tested on standard laptops and/or computers, provided that the connection to the \gls{robovaas} server was available. After following the standard \gls{asv} booking procedure, the companion software (where the low-level controller is running) opens an \gls{rmq} queue and binds it to the exchange where the output of the underwater communication network is broadcast by the emulated sensor node.

To test the signal chain from the underwater sensing equipment, which was hosted in Italy, and communication to the radio network and finally to the online server, hosted in Germany, we developed an in-laboratory test facility.
\begin{figure}[h] 
       \centering
       \subfloat[Phono-absorbent testing tank]{\includegraphics[width=0.4\columnwidth]{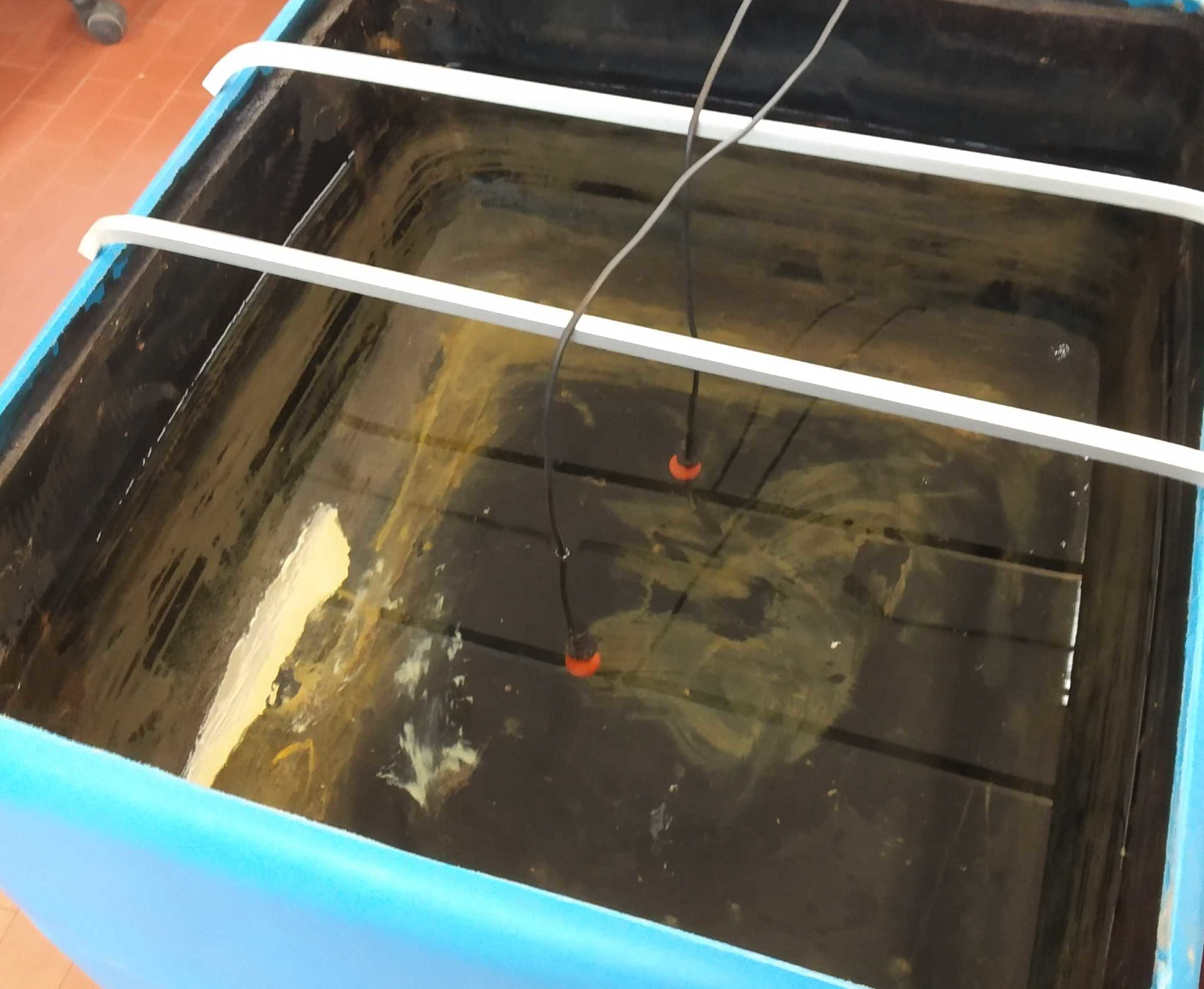}
       \label{fig:ahoi-lab-setup}
       }
       \subfloat[In-lab setup scheme]{\includegraphics[width=0.6\columnwidth]{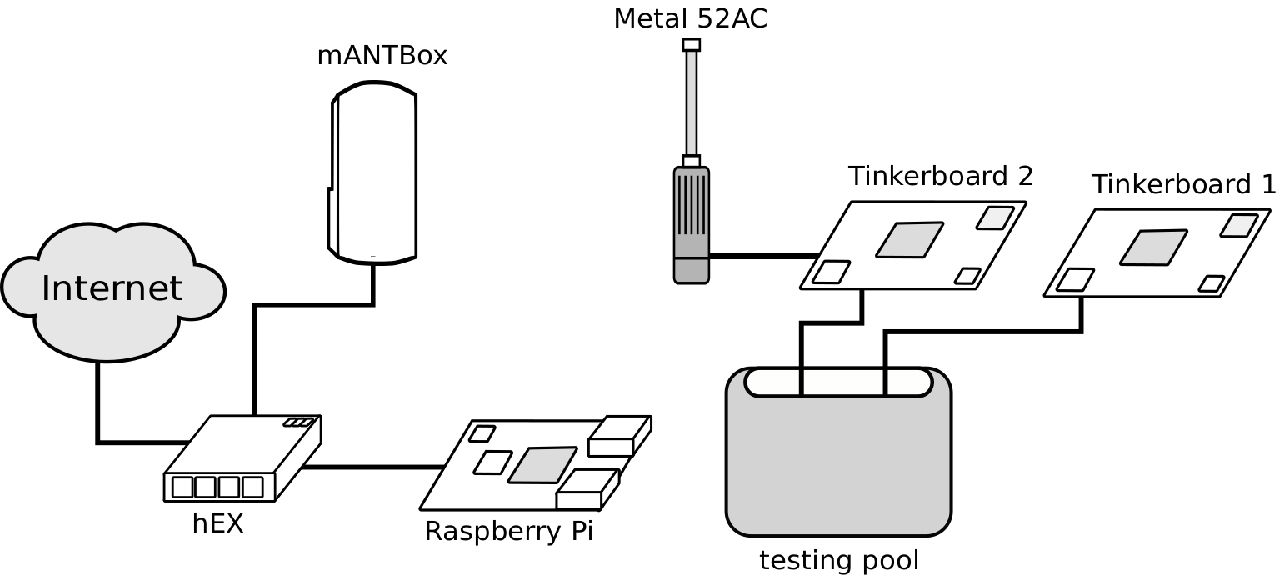}
       \label{fig:signet-seup}
       }
       \caption{\ahoi laboratory setup: while the two underwater nodes and the long range above water network were installed in Padova, the \gls{amqp} server was deployed in Germany and reached through the Internet.}
       \label{fig:test_tank}
\end{figure}
At the SIGNET laboratory (University of Padova) we developed a small tank, with a capacity of about 200 litres, which has sound absorption capabilities through its internal coating~\cite{SIGtank}; this tank does not allow to test topologies but is very useful to test all the hardware and configurations before the sea trial, without stressing the devices with in-air transmission. Figure~\ref{fig:test_tank} shows the setup: the \ahoi transducers inside the tank are presented in Figure~\ref{fig:ahoi-lab-setup}, while the scheme of the tested setup is depicted in Figure~\ref{fig:signet-seup}. The underwater network was composed by two nodes, one emulating a sensor node and one emulating the \gls{asv} communication system (we remark that the \gls{asv} motion and control system, instead, is emulated by the aforementioned digital twin). Both nodes were composed by an \ahoi modem and an ASUS Tinkerboard S \gls{sbc}, where the whole communication framework was running. In addition, the node emulating the \gls{asv} was equipped with a Metal 52AC antenna to transmit the data to the emulated shore station. In this way we could test and debug inside our lab the same equipment and tools used in the trial.

In a first test, the shore station and its \gls{amqp} server were installed in a Raspberry Pi 4 \gls{sbc} directly connected to the emulated \gls{asv} through a WiFi link established connecting the Raspberry Ethernet port to the hEX router that, using the mANTBox antenna, reached the node connected to the Metal 52AC.

In a second test the Raspberry Pi was removed, and the hEX was connected to the Internet so the synthetic data was able to reach the public \gls{robovaas} \gls{rmq} server in Germany. Because the underwater network and the above water link were simulated in Italy, while the Gazebo simulation and the \gls{robovaas} service platform were deployed in Germany, a publicly available \gls{rmq} server was deployed and the underwater network and Gazebo simulations were synchronously started.


On the user and operator end-points there is no need for any change, as the entities of the \gls{robovaas} cloud infrastructure continue to listen to the same \gls{amqp} queues. The \gls{rmq} clients were configured to communicate over the same publicly available \gls{rmq} server, using the predefined queues and message structure. 

\begin{figure}[ht]
    \centering
    \includegraphics[width=\columnwidth]{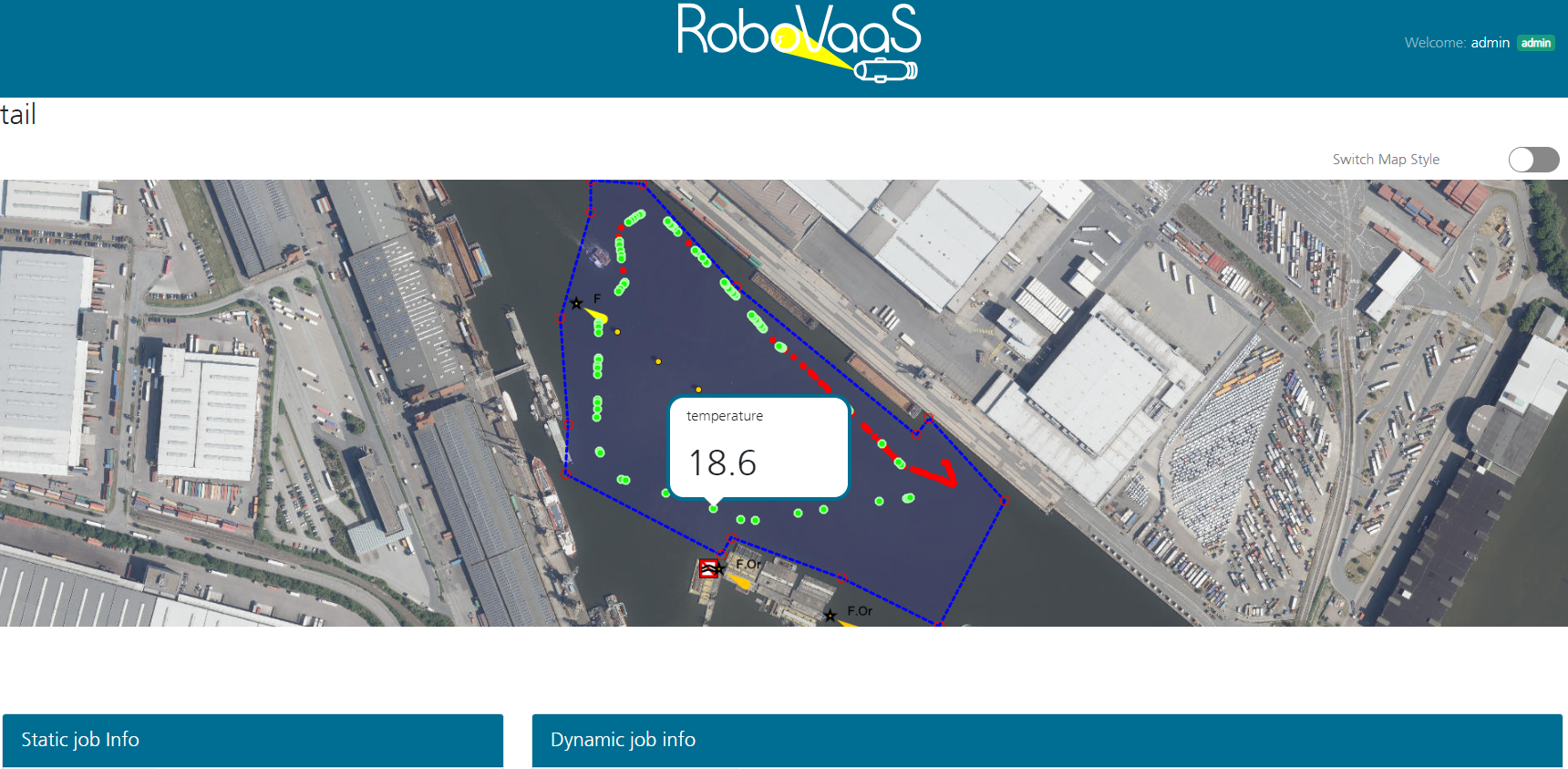}
    \caption{Simulation results}
    \label{fig:simulation_data_muling}
\end{figure}

Figure~\ref{fig:simulation_data_muling} presents the web interface observed by the operator to monitor the mission status: the red markings on the map represent the path followed by the \gls{asv}, each point being a \textit{heartbeat} message successfully received by the server. The green markings are also successfully acknowledged messages but they also indicate the presence of a sensor reading coming from the underwater equipment. 
We underline that from the operator's perspective there is no difference between the emulation and the actual sea trial.

\section{Field Tests: Setup and Exercise Definition} \label{sec:deploy}

Field tests were performed at lake Kreidesee in Hemmoor, Lower-Saxony, Germany during a 10-day campaign. The setup used the same software tools and applications to gather sensor data and send control commands, but distributed over the multiple companion computers and/or laptops. As observed in Figure~\ref{fig:system_architecture}, the onshore equipment included a mobile station acting as a server, supporting the main above water communication brokers and saving the critical mission information into the database. Other mobile stations were used to run the applications that were directly commanding the \glspl{uv}. For the off-shore counterparts, which were previously either simulated or run in confined environments, besides the \gls{asv}, five buoys with ballast to avoid drifting were manually deployed in the different topologies. Regarding the mobile node, the SeaML \gls{asv} carried all the necessary equipment to perform all the services, which varied from a single-beam echo-sounder for the environmental data collection use-case, to the off-shore underwater communication station for the data-muling use-case and, lastly, to a mini-\gls{rov} for the quay walls inspection use-case.

\remembertext{weather}{\fc{The weather conditions were quite stable during the whole test campaign. Specifically, during the day the temperature ranged between 15~$^\circ$C and 25~$^\circ$C and the weather was sunny or cloudy all days. The water temperature at the surface was 19~$^\circ$C and on the lake bottom was 16~$^\circ$C. Almost no wind was observed and there were no waves in the lake. }}

\subsection{Shore Operation Centre}

\begin{figure}[htb] 
       \centering
       \subfloat[OperatorUI App for high-level commanding SeaML \gls{asv}]{\includegraphics[width=0.42\columnwidth]{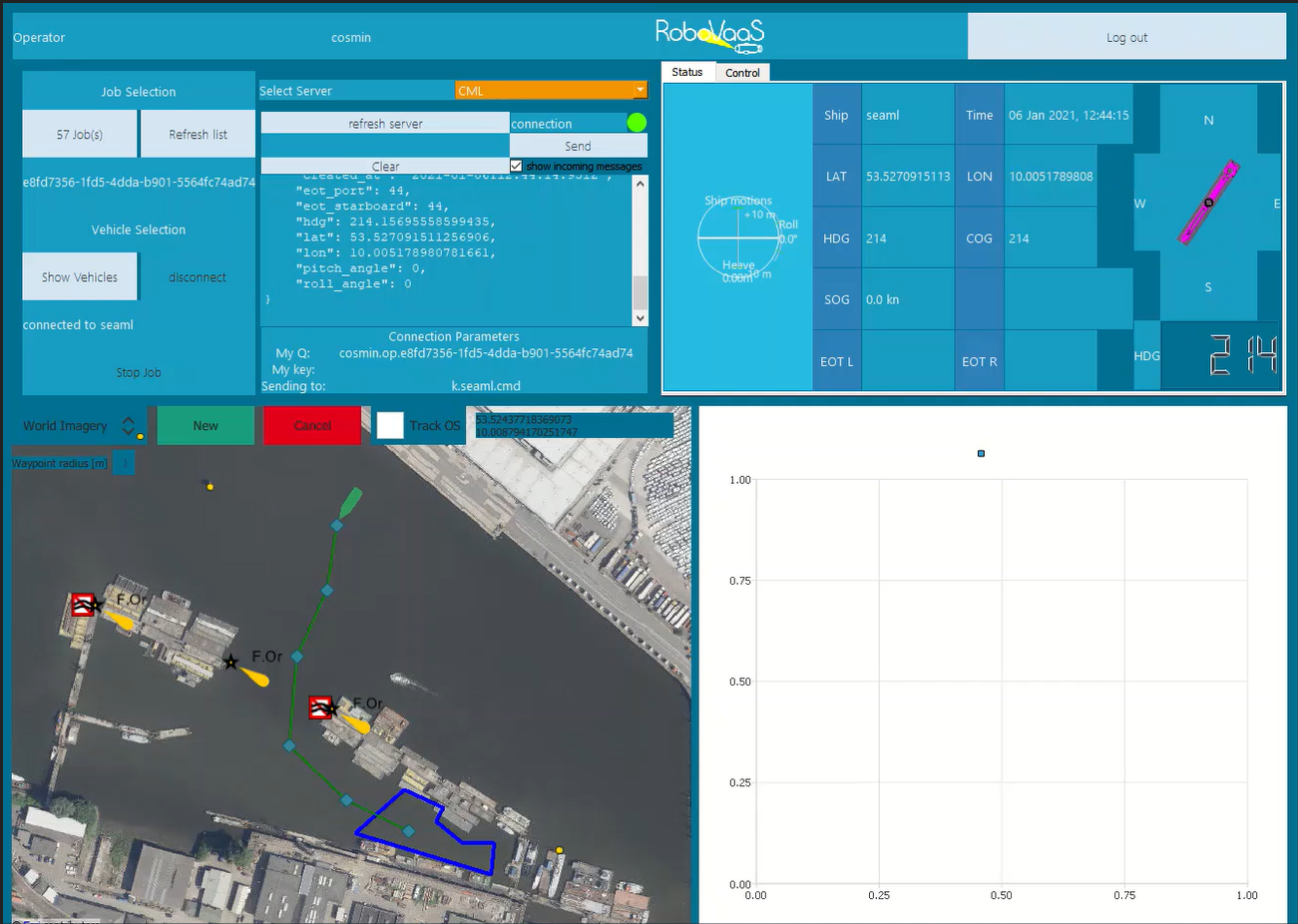}}
       \hspace{0.1 cm}
       \subfloat[QGroundControl App for ROV teleoperation]{\includegraphics[width=0.55\columnwidth]{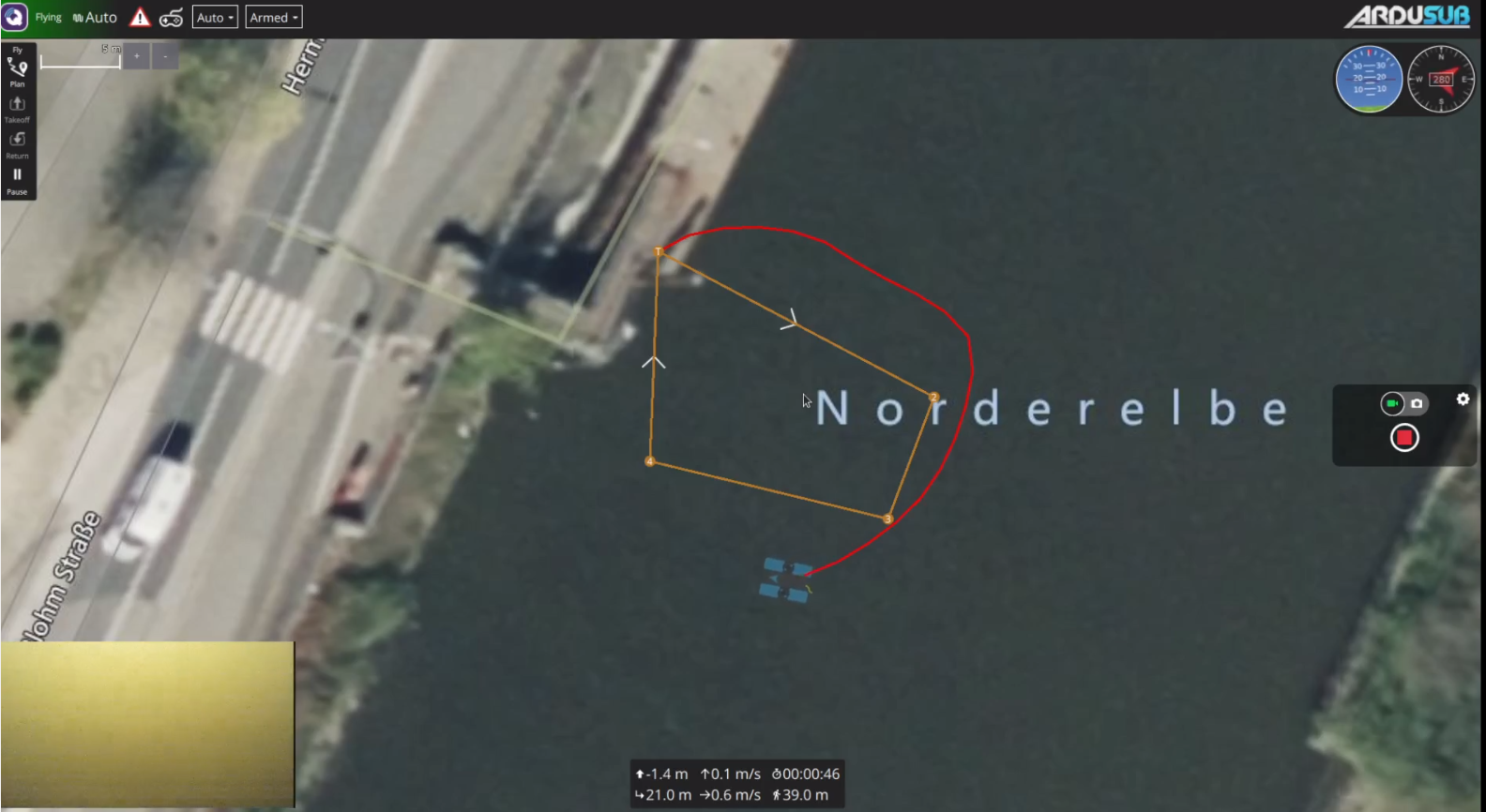}}
       \caption{The Operator Console contains two standalone applications that are used to command and control one \gls{asv} and one \gls{rov}, respectively. The \gls{asv} console is the in-house developed \textit{OperatorUI} while for the \gls{rov} the open-source QGroundControl application is used}
       \label{fig:operator_apps}
\end{figure}

On the shore, the following hardware components were used to deploy the software solution:
\begin{itemize}
    \item the server-side application was deployed on a Dell Latitude 5411 laptop running Ubuntu 18.04;
    \item the operator application, shown in Figure~\ref{fig:operator_apps}, was deployed on a Dell Latitude 7480 laptop running Windows 10;
    \item the underwater network was controlled and monitored from a Panasonic Toughbook CF-53, running Linux Mint 18.1 and connected via SSH to the buoys;
    \item the above water communication network was deployed on a Mikrotik hEX PoE lite, which had a 4G modem attached for internet connectivity and Mikrotik mANTBox 2 12S Wi-Fi antenna.
\end{itemize}

\subsection{Robotic System}

The carrier board was the already mentioned SeaML \gls{asv}, which was designed to fit different measurement equipment and even a mini-\gls{rov}. The absolute maximum payload is \SI{25}{\kilo\gram}, which was never attained during the tests for all use-cases mentioned. The main components of the barebone \gls{asv} are:

\begin{itemize}
    \item 4x T200 propulsion units from BlueRobotics (2 on each side);
    \item 2x Raspberry Pi 4 \glspl{sbc}, one for position estimation, one for computing the command primitives from the high-level commands received from the shore. The first made use of Emlid Navio2 and Reach M+ boards as data acquisition boards and made use of its own low-level controller for fusing sensor data (instead of using an onboard sensor fusion module);
    \item 1x Netgear GS305 switch for connecting all clients to the above water network through the Mikrotik Metal 52AC;
    \item 1x FrSky X8R RC receiver coupled to the propulsion unit for emergency takeovers.
\end{itemize}

Saving a total of nominal \SI{1184}{\watt\hour} for the power transmission and another \SI{148}{\watt\hour} for the data and logic, the SeaML \gls{asv} was able to perform the tasks without battery exchange for \SIrange{8}{12}{\hour}, a time interval which was sufficient to perform \SIrange{4}{8}{} exercises. The obtained autonomy cannot be considered as absolute, because the time intervals also included equipment exchange, calibration and on-site debugging, time in which the propulsion units were either little used or not at all.

All additional components could be connected to the control unit via watertight RJ45 connectors, that forwarded the connections to the (unmanaged) switch. In the next sections, the layout and interfaces of the components specifically aimed for the data-muling use-case are presented.

\subsection{Above Water Network Setup}
\begin{figure}[tb] 
       \centering
       \includegraphics[width=0.7\columnwidth]{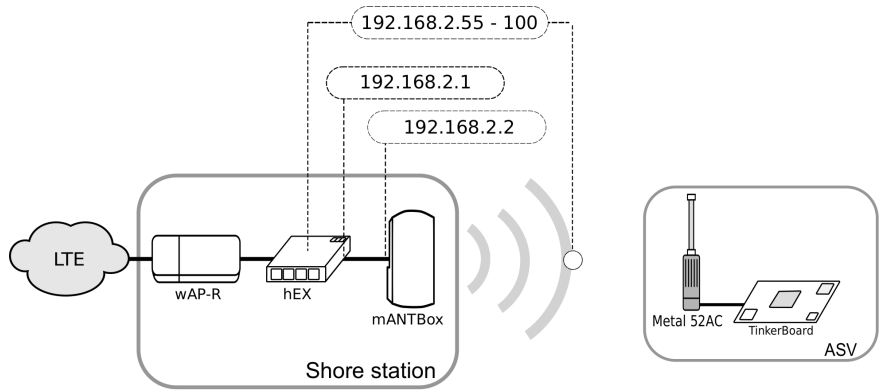}
       \caption{Above water network architecture. The shore station (left), exchanges data with the \gls{asv} (right) using a long range WiFi link, and acts as the gateway of the network providing all nodes Internet connectivity thought LTE.}
       \label{fig:wifi-net-scheme}
\end{figure}
\remembertext{abovewater_depl}{{The underwater communication and all the off-shore devices were supported by a radio network operating in the \SI{2.4}{\GHz} WiFi band and implemented using Mikrotik devices. 
The IP address pool used by the DHCP server was ranging from 192.168.2.55 to 192.168.2.100, while the address of the DHCP server was 192.168.2.1, and the address of the wired interface of the long range shore WiFi antenna was 192.168.2.2: the whole network connections and addressing are depicted in Figure~\ref{fig:wifi-net-scheme}. 
The 12 dBi 120 degree aperture antenna was deployed in the shore at a height of approximately 15~m from the water level, pointing the area where the SeaML ASV was moving. The omni-directional antenna in the ASV was installed on top of the vehicle at a height of approximately 0.75~m. Given the long range requirement, the \SI{20}{\mega\hertz} sub-channels of the 802.11n protocol was used. The effective radiated power (ERP) of the antennas was set to 100~mW, according to the European restrictions~\cite{etsi-2.4GHz}.
A detailed description of the above water network topology and the devices used in the network have been described in Section~\ref{sec::wifinetwork}.
}}

\subsection{Underwater Nodes} 

The underwater nodes for the data collection service are equipped with sensors to measure environmental data, a power supply, a processing unit, and the \ahoi acoustic underwater modem. Usually the nodes are submerged in an area of interest and do not have any cabled connectivity or above water communication link. 
In a preliminary test performed to evaluate the underwater nodes functionality in combination with the \ahoi modems we experienced a communication range between \SIrange{150}{250}{\metre} in static scenarios and up to \SI{150}{\metre} in mobile scenarios.


\begin{figure}[tb] 
       \centering
       \subfloat[Electronic components]{\includegraphics[width=0.32\columnwidth]{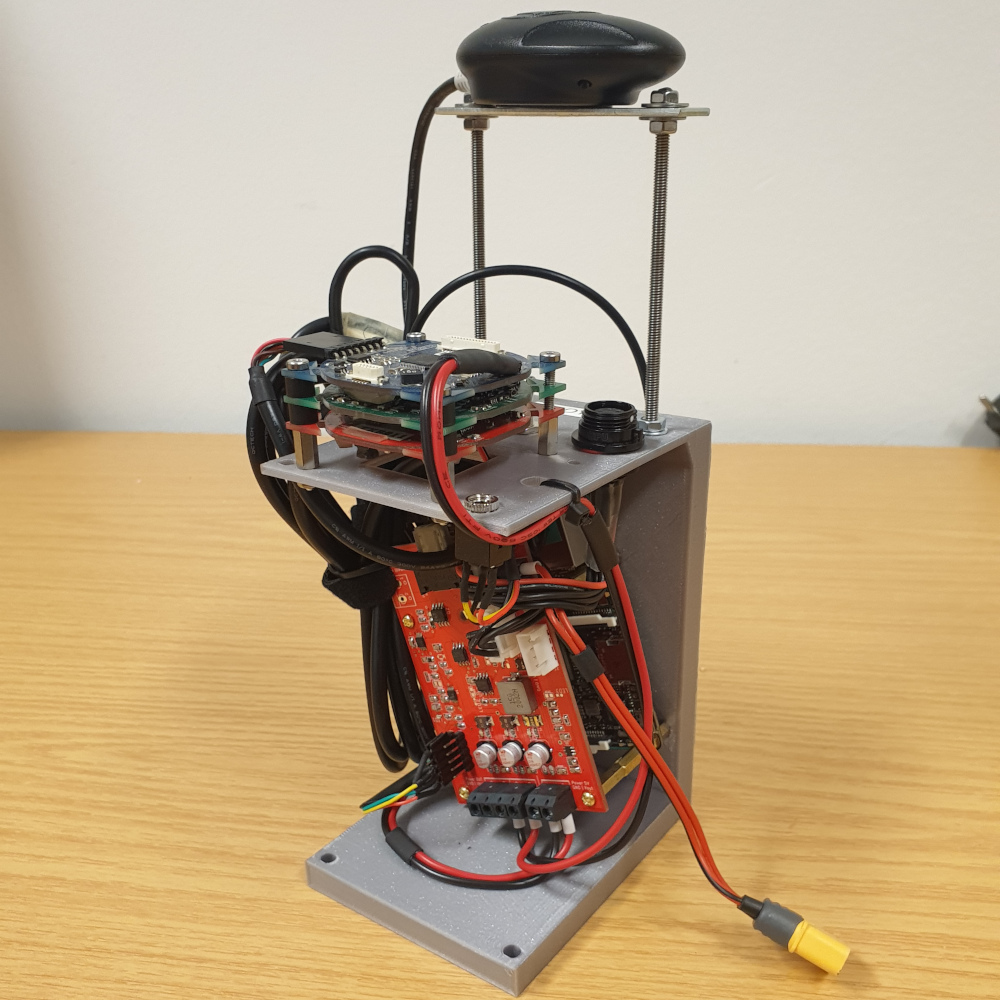}}
       \hspace{0.1 cm}
       \subfloat[Buoys with hydrophones]{\includegraphics[width=0.32\columnwidth]{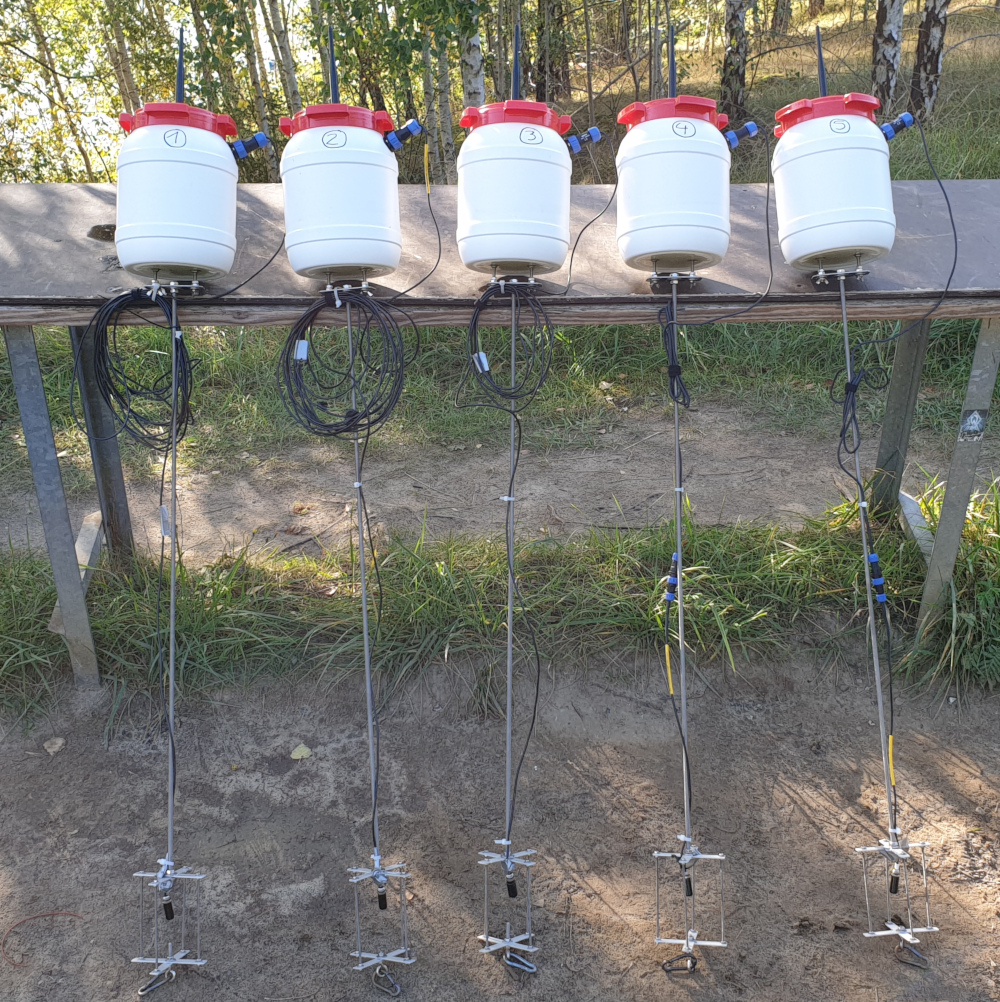}}
       \hspace{0.1 cm}
       \subfloat[Deployed buoy and SeaML]{\includegraphics[width=0.32\columnwidth]{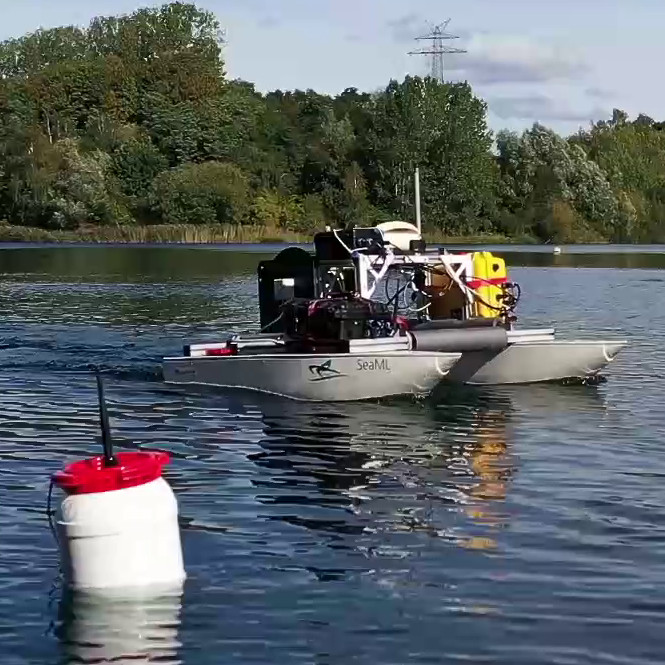}}
       \caption{For the evaluation and demonstration of the data collection service a prototype node based on a buoy was built.}
       \label{fig:buoys}
\end{figure}

To evaluate the data collection service, we decided to use buoys instead of fully submerged nodes to simplify the deployment. A buoy consists of an ASUS TinkerBoard~S used to execute the DESERT Underwater Framework, an \ahoi modem, a Navilock NL-6002U GPS receiver, and a power supply. 
An external WiFi antenna connected to the TinkerBoard enables debugging and monitoring during the evaluation. However, WiFi connection and GPS receiver are just part of the prototype for the service demonstration with buoys, and would be removed from the actual nodes in the case of a long-term deployment. 
For power supply a battery (\SI{11.1}{\volt}, \SI{2400}{\milli\ampere\hour}) and a power management board are used. The power management board provides a stabilized \SI{5}{\volt} supply to the TinkerBoard, measures the battery voltage and current, and protects the components against under-voltage, reverse polarity and short circuits. Depending on the computational load of the TinkerBoard and the transmission intervals of the \ahoi modem, the battery allows an operation time between \SI{6}{\hour} and \SI{10}{\hour}.
The external hydrophone is placed about \SI{1.1}{\metre} under the buoy to avoid the region directly under the water surface that is strongly affected by acoustic reflections. Furthermore, a cage protects the hydrophone against physical damage. Finally, a rope is connected to the protection cage to fix the buoys in the water with an anchor. 

For the deployment, five buoys were constructed. \Cref{fig:buoys} depicts electronic components, the buoys before the deployment and a single buoy with the SeaML \gls{asv}. In addition, a sixth node was prepared, using a Raspberry Pi~4 board connected to an \ahoi modem. The electronic components were installed in a box to place the node on a jetty and submerge the hydrophone at \SI{1}{\metre}~depth.

To collect environmental data from the sensor nodes, the SeaML was equipped with an \ahoi modem and a TinkerBoard as well. Similar electronic components from buoys were installed in a waterproofed case on top of the SeaML. Furthermore, a hydrophone in a protection case was mounted \SI{0.9}{\metre} under the SeaML.
In order to measure the position of the mobile and the fixed node on a jetty, a Navilock NL-8001U GPS receiver was used. 

\subsection{Underwater Network Settings}\label{sec:uw_net_settings}
The underwater network is composed by one mobile node and a number of static nodes that ranges from four to six, depending on the considered network topology. 
In our scenario, every \SI{60}{\second} each static node generates  a packet with a payload of \SI{13}{\byte} (if we consider the packet headers introduced by the communication protocols the size increases up to \SI{25}{\byte}). The transmission bitrate used by the \ahoi modems is \SI{200}{\bit\per\second}.

To collect the data from the sensor nodes, the \textit{UWPOLLING} \gls{mac} protocol has been used. An important parameter of this protocol is the maximum backoff time $T_{b_{max}}$ used in the discovery phase by the sensor nodes to randomize the channel access trying to reduce the collisions (more details are provided in Section~\ref{sec:uwpolling}). In our tests, we set $T_{b_{max}} =$ \SI{15}{\second}. In addition, each node can transmit in each protocol cycle (i.e., every polling phase) a burst of up to five consecutive packets, in order to limit the number of packets sent by a node in each cycle and thus reduce the probability that the \gls{asv} moves out of range during a packet burst transmission.

\subsection{Data-Muling Topologies}\label{sec:muling-topologies}

\begin{figure}[h]
    \begin{subfigure}{.5\textwidth}
        \centering
        \includegraphics[width=.95\linewidth]{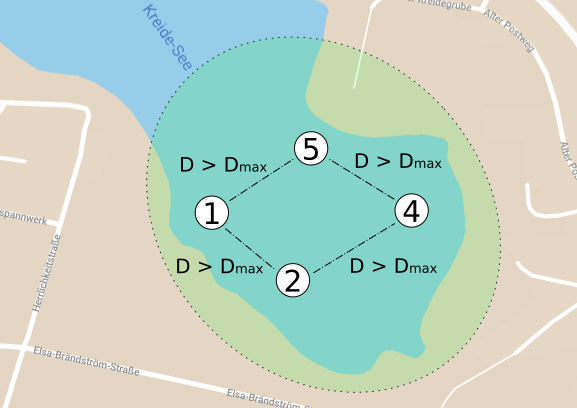}
        \caption{}
        \label{fig:top1_th}
    \end{subfigure}%
    \begin{subfigure}{.5\textwidth}
        \centering
        \includegraphics[width=.95\linewidth]{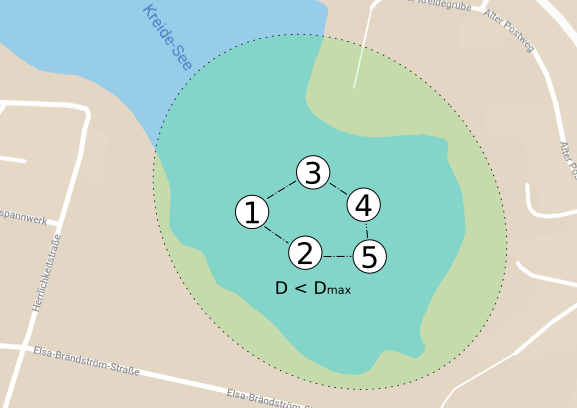}
        \caption{}
        \label{fig:top2_th}
    \end{subfigure}\\
    \begin{subfigure}{.5\textwidth}
        \centering
        \includegraphics[width=.95\linewidth]{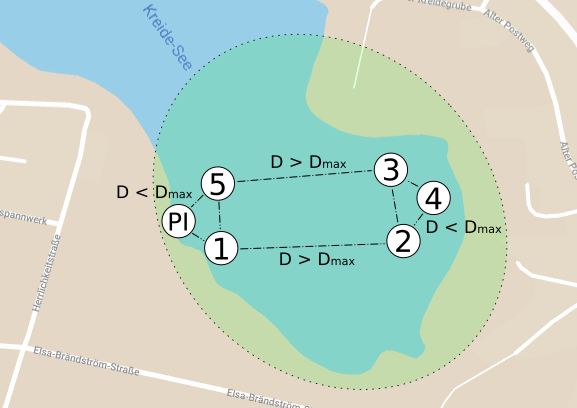}
        \caption{}
        \label{fig:top3_th}
    \end{subfigure}%
    \begin{subfigure}{.5\textwidth}
        \centering
        \includegraphics[width=.95\linewidth]{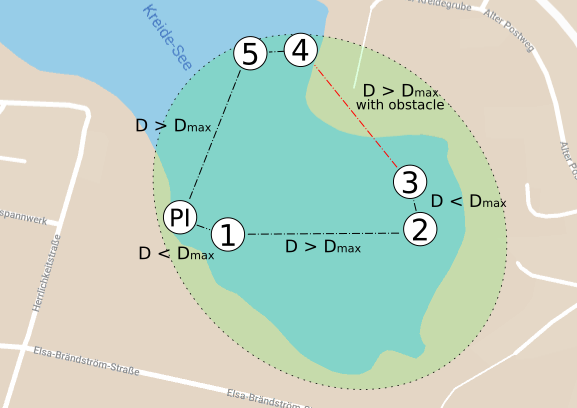}
        \caption{}
        \label{fig:top4_th}
    \end{subfigure}
    \caption{Theoretical topologies represented with the required minimum distances.}
    \label{fig:topALL_th}
\end{figure}
As previously mentioned, the underwater network relied on the DESERT Underwater Framework to accomplish the network protocol management.
The communication is performed with acoustic signals, so that the sensors could be placed even on the seabed and still transmit wirelessly to the surface.
Moreover, the data is transmitted using the DESERT Framework, whose packet header contains the required information for the routing and its packet payload is formatted to include all the needed information for the data acquisition, as explained in Section~\ref{ch:data_compr}.

To better understand the behavior of the \textit{UWPOLLING} \gls{mac} protocol, different network topologies have been tested to analyze how the protocol behaves under different network conditions, such as node density and distances.
The performance has been assessed in terms of fairness and packet delay, and the overall underwater network has been evaluated taking into consideration the \gls{pdr} of the data packets.

To perform this analysis we designed four different topologies, to reproduce four networks with different node densities. In particular, we designed the networks to obtain clusters with a different number of nodes, to test the behavior of the discovery phase of the \textit{UWPOLLING} protocols. Based on the maximum range $D_{MAX}$ of the \ahoi modem (about \SIrange{150}{250}{\metre}) and the area of operations in Figure~\ref{fig:topALL_th}, we depicted the following topologies:


\begin{enumerate}

    \item \textit{equally distanced nodes with $D > D_{MAX}$} : in this topology all the adjacent nodes have a distance greater than the transmission range, thus the \gls{asv} would be in range with no more than two sensor nodes at the same time. This topology is shown in Figure~\ref{fig:top1_th};
    
    \item \textit{equally distanced nodes with $D < D_{MAX}$} : in this topology the nodes are placed close enough to obtain a single cluster with all the available nodes, such that the \gls{asv} can be in range with all the sensor nodes during the entire test. The goal is to let all nodes participate to the same channel contention during a discovery phase. This topology is shown in Figure~\ref{fig:top2_th};
    
    \item \textit{two clusters at $D > D_{MAX}$} : in this topology there are two main clusters, composed of three sensor nodes each and placed at a distance such that when the AUV is close to one cluster it is not in range with the other one. With this topology we obtain a network in which one of the clusters is not visited for a while and accumulates packets in the internal queue until the next visit, when the muling node returns in range. This topology is shown in Figure.~\ref{fig:top3_th};
    
    \item \textit{three clusters at $D > D_{MAX}$ and physical obstacle} : in this topology there are three clusters, composed of two sensor nodes each and placed at a distance larger than the communication range; in addition to the previous case, two clusters are divided also by a physical obstacle (in the Hemmoor test it was a small headland), so that there is a certain communications shielding between the \gls{asv} and the sensor nodes in the other clusters, even while traveling to them. This topology is shown in Figure~\ref{fig:top4_th}.

\end{enumerate}


\section{Field Test: Results} \label{sec:result}

The analysis of the results is covered in this section, considering the underwater and above water networks separately. The performance of the motion controller of the SeaML \gls{asv} is not evaluated, and is considered out of scope. The data-muling exercises are fully dependent on the carrier platform and its ability to integrate the incoming data from the underwater network with those from navigation, and on the flawless correlation of real-time data with the rest of the exercise data contained in the database.
Indeed, the parameters to assess the performance differ since the underwater and above water channels are different by nature, and the protocols are built taking into account the characteristics of the two channels, leading to lower throughput and higher delays for the underwater network.

\subsection{Service Application Performance Analysis}

In the process of collecting environmental data from the underwater nodes, several points that define the vision of \gls{robovaas} have been achieved. First of all, the \gls{asv} has proven to be a reliable platform carrier for all the services we considered. Although within the scope of the project only one \gls{asv} for the small-scale evaluations has been developed and, subsequently, the equipment had to be adapted for each use-case, the changes could be rapidly made thanks to the modular design. 

There were in total 16 complete test runs for the data-muling use-case, where each time the area of interest was selected by a web-browser client and approved by the system broker, while the path computed by the \gls{asv} to patrol the area was finally set by the operator. Physically, the roles of the web-browser client and of the system broker were played by the same person, because both demanded little interaction and access to a web-browser. The operator role was played by another person, who constantly monitored the job execution and focused on avoiding collisions or loss of signal. In case of emergency, the operator was equipped with an RC remote, that could take over control from the autopilot and sail the \gls{asv} by manually steering it. This was however seldom used, because the above water network proved to be very reliable even around corners where some natural elements, such as trees, were blocking the direct contact to the \gls{asv}. Secondly, the chosen location had no ship traffic so most of the obstacles where static, which allowed precise path planning. Each time the operator chose the appropriate path for the \gls{asv} to follow, so that all sensor buoys were slalomed before it returned to shore. 
In order to evaluate the performance of the system in a mission that is as close as possible to reality, the path was changed even for the same topology; the distances from the \gls{asv} and the sensor buoys were 
always longer than \SI{2}{\metre} in order to avoid collisions due to sway. 

\begin{figure}
    \centering
    \includegraphics[width=\columnwidth]{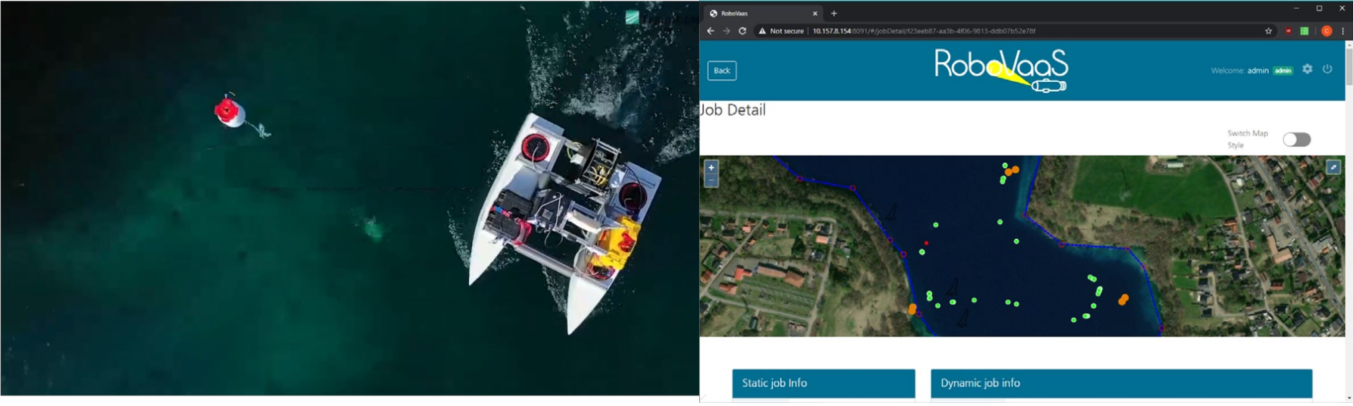}
    \caption{Underwater data collection service results visualization obtained in real time from the messages acquired by the \gls{asv} from the underwater wireless nodes and sent to the \gls{rmq} queues generated for this exercise. Points without sensor data, previously marked with red dots, have been removed from visualization}
    \label{fig:experiment_data_muling_webui}
\end{figure}

The average speed of the \gls{asv} was around \SI{1.5}{knots} (\SI{0.78}{\metre\per\second}) and the paths ranged from \SI{700}{\metre} to \SI{1.2}{\kilo\metre}. Due to the fact that a waypoint following algorithm was used and because the catamaran-shaped \gls{asv} is an underactuated system, the paths could not be perfectly followed. The effects of velocity perpendicular to the direction of motion (known as sway) increased the total distance followed, but this additional distance was hence neglected. As observed in Figure~\ref{fig:experiment_data_muling_webui}, the packets from the underwater nodes have been constantly collected during the exercise, with increasing frequencies when the \gls{asv} was within a \SI{60}{\metre} radius from the nodes.

\begin{figure}
    \centering
    \includegraphics[width=\columnwidth]{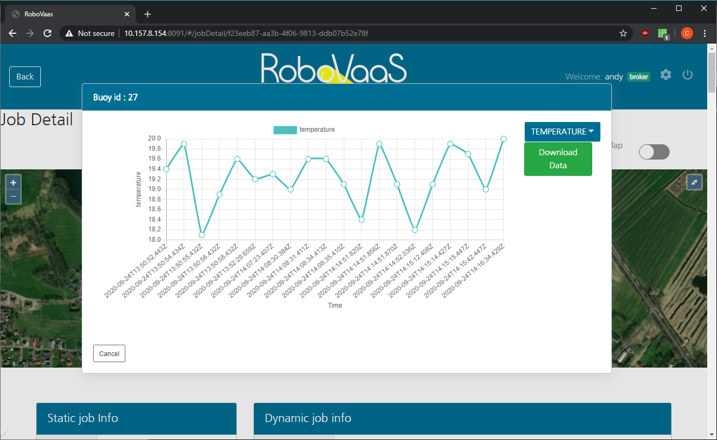}
    \caption{Graphical representation of underwater data collection service on the web user interface. Visualization generated online for each buoy and updated in real time.}
    \label{fig:experiment_data_muling_buoy_stats}
\end{figure}

Each packet of data received from the underwater network was incorporated in the messages delivered to the shore over \gls{amqp} via \gls{rmq}. The \gls{asv} incorporated the JSON-encoded messages into a JSON object containing the keyword \textit{data}. The front-end server handling the web user interface of the web-browser client then handled the object as a sensor point and marked it accordingly on the map shown to the user. Secondly, the contents being saved by the web application back-end into the database with the appropriate exercise-specific identification number, further metrics could be shown, as depicted in Figure~\ref{fig:experiment_data_muling_buoy_stats}. Lastly, the web-browser clients could download the plotted data in \gls{csv} format for further analysis.

\subsection{Above water Network: Performance Analysis}

An evaluation of the above water network was taken separately during the last days of tests and were performed, at first, considering only the actual coverage of the mANTBox~Access~Point. 
We recall that the \SI{2.4}{\giga\hertz} WiFi channel was used with a \SI{20}{\mega\hertz} bandwidth and the transmission power was configured to the maximum allowed by the European WiFi regulations (\SI{100}{\milli\watt})~\cite{etsi-2.4GHz}.
To evaluate the WiFi coverage we set a continuous PING command from the SeaML \gls{asv} to the mANTBox and manually drove the SeaML in different places of the lake and saved the data in local log files.

\begin{figure}[h]
    \centering
    \includegraphics[width=0.55\linewidth]{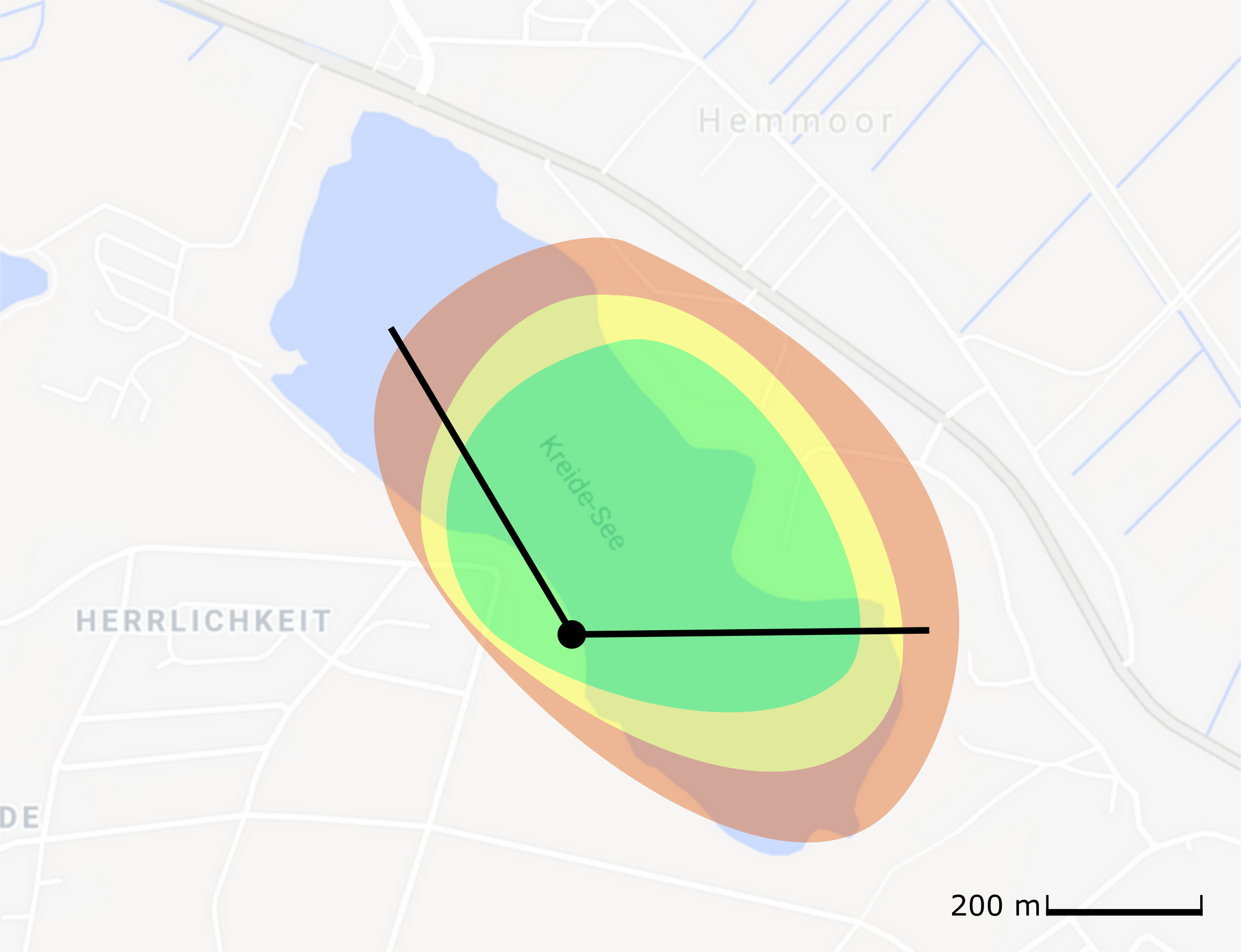}
    \caption{Coverage map of mANTBox WiFi over Hemmoor Lake. A stable connection was achieved up to a maximum range of \SI{350}{\metre} from the shore station.}
    \label{fig:coverage}
\end{figure}

We use iperf3~\cite{iperf} software to measure the performance of the network. Specifically, we use iperf3 to transmit a TCP and a UDP stream between the two antennas and measure the maximum achievable throughput, computed as the number of payload bits received at the application layer per second, and the jitter, that provides the variation of the packet delivery delay, and can be computed as~\cite{rfc3393} 
\begin{equation}
     jitter = \sum_{i=1}^{N_v-1} \frac{|d(i+1) - d(i)|}{N_v-1},
\end{equation}
where $N_v$ is the number of packets received for the considered video stream, and $d(i)$ is the delay of the $i$-th packet. 
This metric is very important for some applications such as video streaming: specifically, a video is considered to be smooth enough to remotely control an unmanned vehicle when the packet jitter is below \SI{2}{\milli\second}~\cite{pdv_bound}.

\remembertext{coverage-zones}{\fc{The obtained outcome is shown in Figure~\ref{fig:coverage}, where different zones are defined based on the delay/jitter retrieved by the PING command with respect to the positions of the \gls{asv}. The antenna is \SI{120}{\degree} directional, but includes secondary lobes which permit transmissions at a wider aperture.
The green area represents the zone where the link is stable, and expands up to a range of approximately \SI{350}{\metre}. In the yellow area the link is unstable, while in the red area the link is disrupted. Specifically, the average throughput available in the green zone is \SI{19.1}{\mega\bit\per\second} and the jitter measured with iperf3 is less than \SI{0.3}{\milli\second}, and so it is small enough to permit a real-time stream video with standard resolution.
Instead, in the border zones (yellow and red), the average throughput available is \SI{17.3}{\mega\bit\per\second} but the connection is really unstable with many disruptions and the jitter oscillates between \SI{0.4}{\milli\second}, when close to the green area, and \SI{1.9}{\second}, when the link is disrupted and then reestablished, depending on the position of the mobile antenna, which means, in \gls{qoe} terms, that several packets could be lost during the transmission and a video stream would be laggy. Furthermore, in the red zone the connection appears discontinuous, which leads to long periods where the devices try to establish a new connection.}}

\remembertext{radio-regulations}{\fc{In terms of coverage, which depends
on the transmission power, the RoboVaaS scenario is restricted by the European ETSI regulation~\cite{etsi-2.4GHz} that imposes a maximum ERP = 100~mW. With this configuration, the use of a \SI{40}{\mega\hertz} channel, that provides a higher throughput\footnote{We measured a throughput of \SI{50}{\mega\bit\per\second} with the antennas placed at a distance of up to 100~m between each other.}, did not allow us to cover the required range for RoboVaaS (at least a few hundreds of meters). Therefore, due to the throughput requirements of less than 10~Mbit/s (Section~\ref{sec:use_case}), the whole evaluation was performed only with the \SI{20}{\mega\hertz} channel that matched the RoboVaaS needs.}}

\subsection{Underwater Physical Layer: Performance Analysis}\label{subsec:res_ahoi}

Before the underwater network trials with the DESERT Underwater Framework, preliminary tests were done to analyze the physical transmission with the \ahoi modems. During the tests, all modems were connected via WiFi to a central computer. The central computer saves all received acoustic packets and triggers the packet transmissions. In all cases, packets with an overall size of \SI{6}{\byte} each 
were used. Previous evaluations pointed out that the bottleneck of \ahoi modems is the synchronization phase at the beginning of the data transmission~\cite{SMP_OCEANS2019_ModemResilience}. Based on that, short packets were used to analyze the communication links.

\begin{figure}[tb] 
       \centering
       \subfloat[SeaML (transmitter) $\Rightarrow$ static nodes (receiver)]
                      {\includegraphics[]{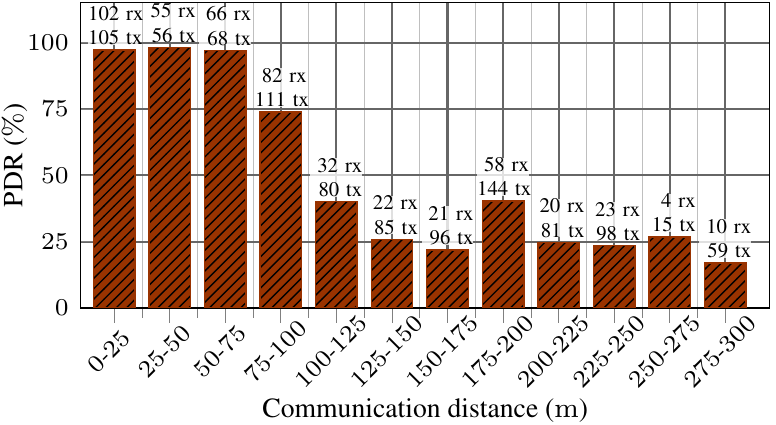}
       				   \label{fig:pdr_seaml_to_buoy}}
       \subfloat[Static nodes (transmitter) $\Rightarrow$ SeaML (receiver)]
                      {\includegraphics[]{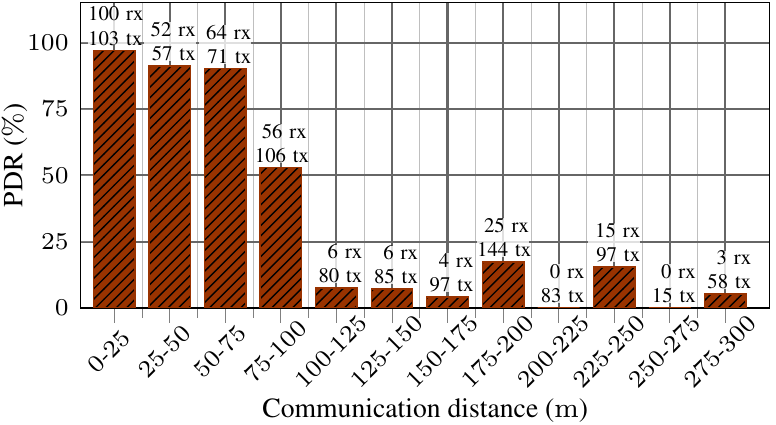}
                       \label{fig:pdr_buoy_to_seaml}}
       \caption{\glspl{pdr} between a mobile node (SeaML) and static nodes (five buoys and node on a jetty). }
       \label{fig:pdr_phy}
\end{figure}

The involved modems transmit in a loop. Every modem transmits a single packet and each following transmission starts \SI{500}{\milli\second} after the end of the previous transmission due to the long propagation time of acoustic signals and long channel delay spread. For each data-muling topology (see \Cref{sec:muling-topologies}), the transmission links between the static nodes are tested. 
During the evaluation, high \glspl{pdr} for ranges up to \SI{150}{\metre} are observed. On average, the \gls{pdr} is \SI{90.1}{\percent} (3605 received packets out of 4000) in this region.
Also for longer static communication links up to \SI{350}{\metre} a high reliability is measured, e.g., \SI{66.0}{\percent} (2377 received packets out of 3600) for communication distances between \SI{300}{\metre} and \SI{375}{\metre}. However, for long distances ($\geq \SI{150}{\metre}$) the \gls{pdr} has a strong dependence on position. At some positions, \glspl{pdr} below \SI{15}{\percent} are observed.

Furthermore, the communication link between static nodes (buoys and node on the jetty) and the mobile node (SeaML) is analyzed. For the test, the SeaML travels a route of \SI{743}{\metre} in the area the static nodes. The trial takes \SI{16}{\minute} and each node transmits 171~acoustic packets with \SI{6}{\byte}~header, \SI{0}{\byte}~payload and \SI{500}{\milli\second}~delay after each transmission. In sum, \SI{1197} packets are transmitted. 
\Cref{fig:pdr_phy} displays the \gls{pdr} between SeaML and static nodes for different distances. The numbers on top of the bars indicate the number of received~(rx) and transmitted~(tx) packets. 
Static communication links between the static nodes are not counted in the bar plot.

The communication from SeaML to the static nodes is better than the communication from the nodes to the SeaML. In both cases, the \glspl{pdr} for distances from~\SIrange{0}{75}{\metre} are between~\SI{90}{\percent} and \SI{98}{\percent}. Conversely, the \gls{pdr} for distances from~\SIrange{75}{100}{\metre} is \SI{74}{\percent} in the first case (SeaML to static nodes) and~\SI{53}{\percent} in the second case (static nodes to SeaML). For longer distances from~\SIrange{100}{300}{\metre} the \glspl{pdr} are between~\SI{17}{\percent} and \SI{40}{\percent} (SeaML to static nodes) and between~\SI{0}{\percent} and \SI{17}{\percent} (static nodes to SeaML). 

In summary, the communication range in static scenarios is longer than in the mobile case. The relative velocity between transmitter and receiver produces a frequency shift at the receiver side (Doppler effect) and a synchronization error. The synchronization error increases with the packet duration.

\begin{figure}[tb] 
       \centering
       \subfloat[Stopped SeaML (without thruster noise)]
                      {\includegraphics[]{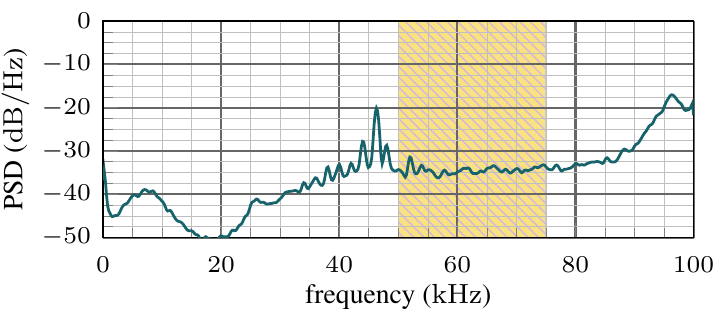}
       				   \label{fig:psd_noisefree}}
       \hspace{0.3 cm}
       \subfloat[Driving SeaML (with thruster noise)]
                      {\includegraphics[]{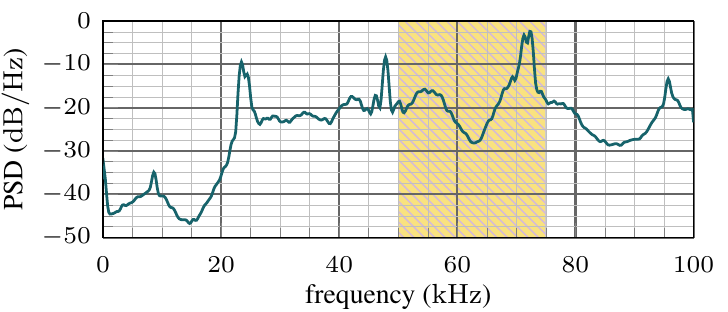}
                       \label{fig:psd_seaml}}
       \\
       \subfloat[Stopped SeaML and communication (\SI{70}{\metre} dist.)]
                      {\includegraphics[]{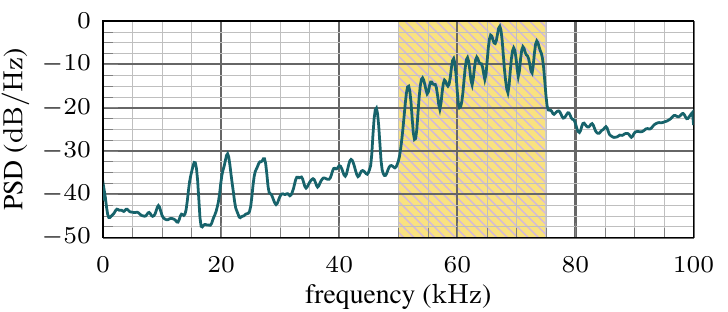}
                       \label{fig:psd_noisefree_pkt}}
       \hspace{0.3 cm}
       \subfloat[Driving SeaML and communication (\SI{68}{\metre} dist.)]
                      {\includegraphics[]{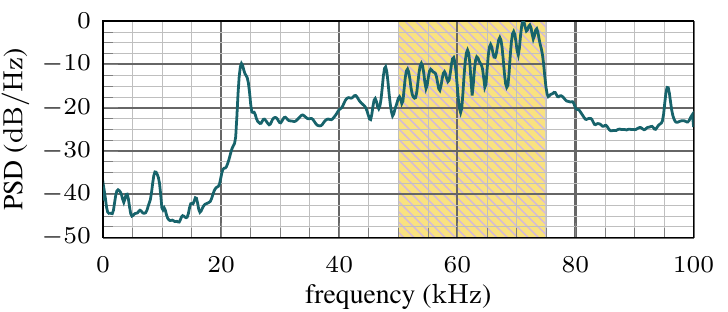}
                       \label{fig:psd_seaml_pkt}}
       \caption{Thruster noise and communication sampled by the \ahoi modem. The yellow area indicates the transmission bandwidth and the plots are normalized to a maximum of~\SI{0}{\decibel\per\hertz}.}
       \label{fig:psd}
\end{figure}

\remembertext{asymmetry}{\fc{The reason for this asymmetrical behavior in the mobile scenario is the acoustic noise produced by the thrusters of the SeaML, hence the transducer on-board the vehicle experienced a higher noise than the transducers deployed in the static nodes.}} 
The SeaML uses four BlueRobotics T200 thrusters, which are mounted on the left and right sides of the SeaML at a distance of about~\SI{1.1}{\metre} from the hydrophone.
\Cref{fig:psd} depicts the received \glspl{psd} for different scenarios. In all cases the \ahoi modem recorded the signal over~\SI{163.84}{\milli\second} (32768~samples with \SI{200}{\kilo\hertz} sampling rate) with a fixed gain. It is important to note that the signal is recorded at the \gls{adc} after the analog signal processing chain. Based on that, the signal is affected by the hydrophone characteristic, a pre-amplifier, a band-pass filter and a second amplifier. For more information about the analog signal processing, the reader is referred to~\cite{renner_tosn}.
At first, the received signal with disabled thrusters is measured (see \Cref{fig:psd_noisefree}). The yellow area indicates the transmission bandwidth between \SI{50}{\kilo\hertz} and \SI{75}{\kilo\hertz}. All analog components are tuned to provide a flat transmission characteristic and a low noise spectrum in the transmission bandwidth. Therefore, other regions are not optimized due to the low-cost and low-power requirements. 
Afterwards, the second diagram shows the recorded signal while the SeaML is traveling at maximum speed (during the experiments). \Cref{fig:psd_seaml} depicts a higher noise level compared to the stopped SeaML. In the communication bandwidth from \SIrange{50}{75}{\kilo\hertz} the average noise level increases from~\SI{-34.5}{\decibel\per\hertz} to~\SI{-18.2}{\decibel\per\hertz}. In addition, the thruster noise produces a large peak around~\SI{72}{\kilo\hertz}. 
The acoustic noise from the thrusters decreases the \gls{snr} between acoustic communication signal and background noise. Therefore, the communication reliability from the SeaML to the static nodes (without thruster noise at the receiver side) is better than the communication from the static nodes to the SeaML. 
For the sake of completeness, \Cref{fig:psd_noisefree_pkt,fig:psd_seaml_pkt} represent \gls{psd} during a packet reception without and with thruster noise. The communication distances are~\SI{70}{\metre} and \SI{68}{\metre} and the average sound level in the communication bandwidth is~\SI{-12.8}{\decibel\per\hertz} (without thruster noise) and~\SI{-10.3}{\decibel\per\hertz} (with thruster noise). In this example the \gls{snr} (w.r.t. the average values in the communication bandwidth) is \SI{21.7}{\decibel} without thruster and \SI{7.1}{\decibel} with thruster noise. The communication distance of about~\SI{70}{\metre} is still in an area with high \gls{pdr} (see \Cref{fig:pdr_buoy_to_seaml}). However, longer communication distances would result in lower \gls{snr} and therefore a lower \gls{pdr}.
\remembertext{other_phy_params}{\fc{Other conditions that usually impact the acoustic communications, such as transducers depth, motion, changes in wind speed and water salinity can be excluded. The depth of the transducers was fixed to approximately \SI{1}{\metre} below the surface thanks to the metal frame used for the deployment. Motion-based Doppler Effects affected the communication link in both directions (static nodes to \gls{asv} and \gls{asv} to static nodes) and does not explain the asymmetric behavior. In addition, wind and waves were rarely noted and constant in the evaluation area. In addition, water temperature and salinity were measured with a professional CTD-48 probe manufactured by Sea\&Sun Technologies near to the node PI (see~\cref{fig:topALL_th}). At 1~m depth, the water temperature was~\SI{18.1}{\celsius} and the salinity was~\SI{0.24}{\ppt}, resulting in a speed of sound of~\SI{1476.6}{\metre\per\second}. Based on the narrow evaluation area and small fluctuations (no water inflow or outflow), temperature, salinity and speed of sound can be assumed to be constant and to affect the communication link in a symmetric way. }}

\subsection{Underwater Network: Performance Analysis}

The underwater scenario is characterized by high propagation delay and bit error rate, combined with a low bitrate determined by the frequencies available for communications. Thus, on one hand, the overall throughput is orders of magnitude lower than that observed in wireless terrestrial networks. On the other hand, the \gls{pdd}, defined as the time elapsed from the packet generation to the correct packet reception at the destination, in underwater acoustic networks is in the order of several seconds or minutes, not of milliseconds like in terrestrial networks.
These phenomena, typical of~\glspl{dtn} such as satellite and underwater acoustic networks~\cite{dtn}, are even more evident in a data-muling scenario, where a mobile node collects the data from submerged sensors when it enters their communication range, as the link between a sensor and the mobile node is systematically disrupted as soon as the mobile node moves out of the sensor's coverage area. Another important metric that shows how the network worked during the tests is Jain's \textit{fairness} index (JFI) \cite{jfi_book}, which indicates whether or not all the nodes were treated fairly by the network protocol stack. Specifically, Jain's Fairness index is computed as 
\begin{equation}
	JFI = \frac{\bigg(\sum\limits_{i=1}^{N_{nodes}}P_{rx,i}\bigg)^2}{N_{nodes} \sum\limits_{i=1}^{N_{nodes}} P_{rx,i}^2} \; ,
\end{equation}
where $N_{nodes}$ is the number of nodes in the network and $P_{rx,i}$ is the number of data packets received by the \gls{asv} from node $i$.

In Tables~\ref{tab:met1}, \ref{tab:met2}, \ref{tab:met3} and \ref{tab:met4}, we present the network performance for the four different topologies.

\begin{table}[h]
\parbox{.45\linewidth}{
    \centering
    \caption{Metrics for Topology 1}
    \label{tab:met1}
\vspace*{-12pt}
    \begin{tabular}{@{}lcccc@{}}
    \toprule
    \textbf{Topology 1} & Sent & Received & PDR {[}\%{]} & \gls{pdd} {[}s{]} \\ \midrule
    Node 1     & 48   & 38       & 0.791                      & 501.46        \\
    Node 2     & 63   & 55       & 0.873                      & 326.55        \\
    Node 4     & 65   & 60       & 0.923                      & 438.97        \\
    Node 5     & 70   & 58       & 0.828                      & 417.30        \\ \bottomrule
    \end{tabular}
}
\hfill
\parbox{.45\linewidth}{
    \centering
    \caption{Metrics for Topology 2}
    \vspace*{-12pt}
    \begin{tabular}{@{}lcccc@{}}
    \toprule
    \textbf{Topology 2} & Sent & Received & PDR {[}\%{]} & \gls{pdd} {[}s{]} \\ \midrule
    Node 1              & 60   & 48       & 0.8                      & 81.16         \\
    Node 2              & 59   & 39       & 0.661                    & 74.49         \\
    Node 3              & 60   & 48       & 0.8                      & 54.7          \\
    Node 4              & 60   & 37       & 0.616                    & 81.08         \\
    Node 5              & 59   & 46       & 0.779                    & 74.41         \\ \bottomrule
    \end{tabular}
    \label{tab:met2}
}

\end{table}

\begin{table}[h]
\parbox{.45\linewidth}{
    \centering
    \caption{Metrics for Topology 3}
    \vspace*{-12pt}
    \begin{tabular}{@{}lcccc@{}}
    \toprule
    \textbf{Topology 3} & Sent & Received & PDR {[}\%{]} & \gls{pdd} {[}s{]} \\ \midrule
    Node 1              & 57   & 42       & 0.737                      & 331.39        \\
    Node 2              & 59   & 46       & 0.779                      & 187.57        \\
    Node 3              & 60   & 34       & 0.566                      & 97.42         \\
    Node 4              & 60   & 44       & 0.733                      & 237.95        \\
    Node 5              & 59   & 37       & 0.616                      & 296.81        \\
    Node PI             & 60   & 32       & 0.533                      & 296.81        \\ \bottomrule
    \end{tabular}
\label{tab:met3}
}
\hfill
\parbox{.45\linewidth}{
    \centering
    \caption{Metrics for Topology 4}
    \vspace*{-12pt}
    \begin{tabular}{@{}lcccc@{}}
    \toprule
    \textbf{Topology 4} & Sent & Received & PDR {[}\%{]} & \gls{pdd} {[}s{]} \\ \midrule
    Node 1              & 50   & 28       & 0.560                      & 233.89        \\
    Node 2              & 50   & 41       & 0.820                      & 571.40        \\
    Node 3              & 60   & 37       & 0.616                      & 267.01        \\
    Node 4              & 57   & 38       & 0.666                      & 438.29        \\
    Node 5              & 56   & 41       & 0.732                      & 255.55        \\
    Node PI             & 51   & 37       & 0.725                      & 255.55        \\ \bottomrule
    \end{tabular}
\label{tab:met4}
}

\end{table}

As defined above, the \gls{pdd} is the delay after which a packet is correctly received, and is computed starting from the packet generation time. Thus, for the topologies where the nodes are not always in the \gls{asv} range, the delay also includes the time the vehicle takes to visit all the other nodes and to come back. This can be observed for topologies 1, 3, and 4, where the average \gls{pdd}s of the networks (i.e., the \gls{pdd} averaged over the nodes) are 421.07~s, 241.33~s, and 336.95~s, respectively,  against the average network \gls{pdd} of topology 2 equal to 73.17~s.
Indeed, in Topology 2 the mean delays per node are far smaller than in the other topologies, because all nodes are in range of the \gls{asv} approximately during the whole test, thus the reception time for the packets is determined mainly by the time the \textit{UWPOLLING} protocol takes to poll every node that successfully takes part in the discovery phase.


From the analysis of the log files obtained during the test campaign, we observe that the \glspl{pdr} change for the different topologies and depend on both preamble synchronization problems (as stated in Section~\ref{subsec:res_ahoi}) and asymmetry of the acoustic link due to the \gls{asv} movement and the noise produced by its thrusters.
Indeed, in some cases a few packets are lost inside the burst of 5 packets sent by a node in its polling phase, while in other cases the burst of packets is completely (or almost completely) lost. In the last case, most of the control packets exchanged by the vehicles and the nodes (i.e., TRIGGER, PROBE, and POLL) are correctly received, but then the link quality drops faster due to the movement and the asymmetry mainly caused by the \gls{asv} propellers' noise, therefore the data packets are lost. While in the first case packet loss happens independently in one or more packets inside the burst of 5 data packets, in the second case the whole burst of 5 packets (or at least the last part of a burst) is lost.
Depending on the topologies and on the specific node position the prevalent cause of packet loss can be different.
In Topology~1, the \gls{pdr} at the \gls{mac} layer (i.e., without considering the packets that remained in the queue at the end of the simulation) is higher than 80\% for all the nodes. In this case, except for node 5, the main cause of packet loss is a failure in the preamble synchronization.
Even in Topology~2, the main cause of packet loss is the failure in the synchronization. Indeed, in this case the \gls{asv} is in range with all the nodes most of the time, and therefore the \gls{asv} movement does not have much of an impact on the reception rate.
On the other hand, the channel asymmetry and the \gls{asv} movement become the prevalent cause of packet loss in Topology~3 and 4, where the \gls{asv} moves between clusters of nodes that are not within range of each other or are even separated by a physical obstacle (as in Topology 4).
\begin{figure}[h]
    \centering
    \begin{subfigure}{.5\textwidth}
        \centering
        \includegraphics[width=.95\linewidth]{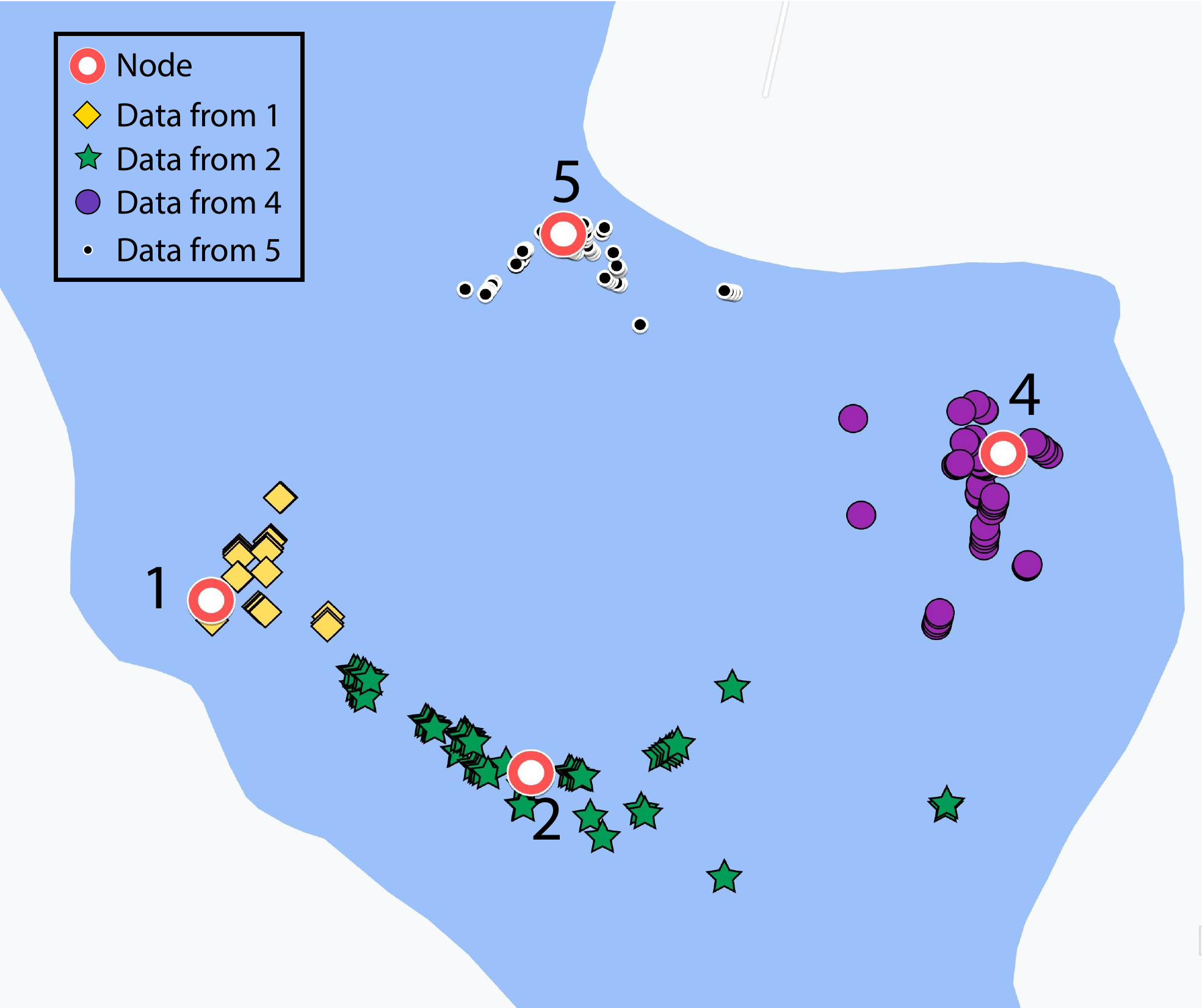}
        \caption{}
        \label{fig:top1_res}
    \end{subfigure}%
    \begin{subfigure}{.5\textwidth}
        \centering
        \includegraphics[width=.95\linewidth]{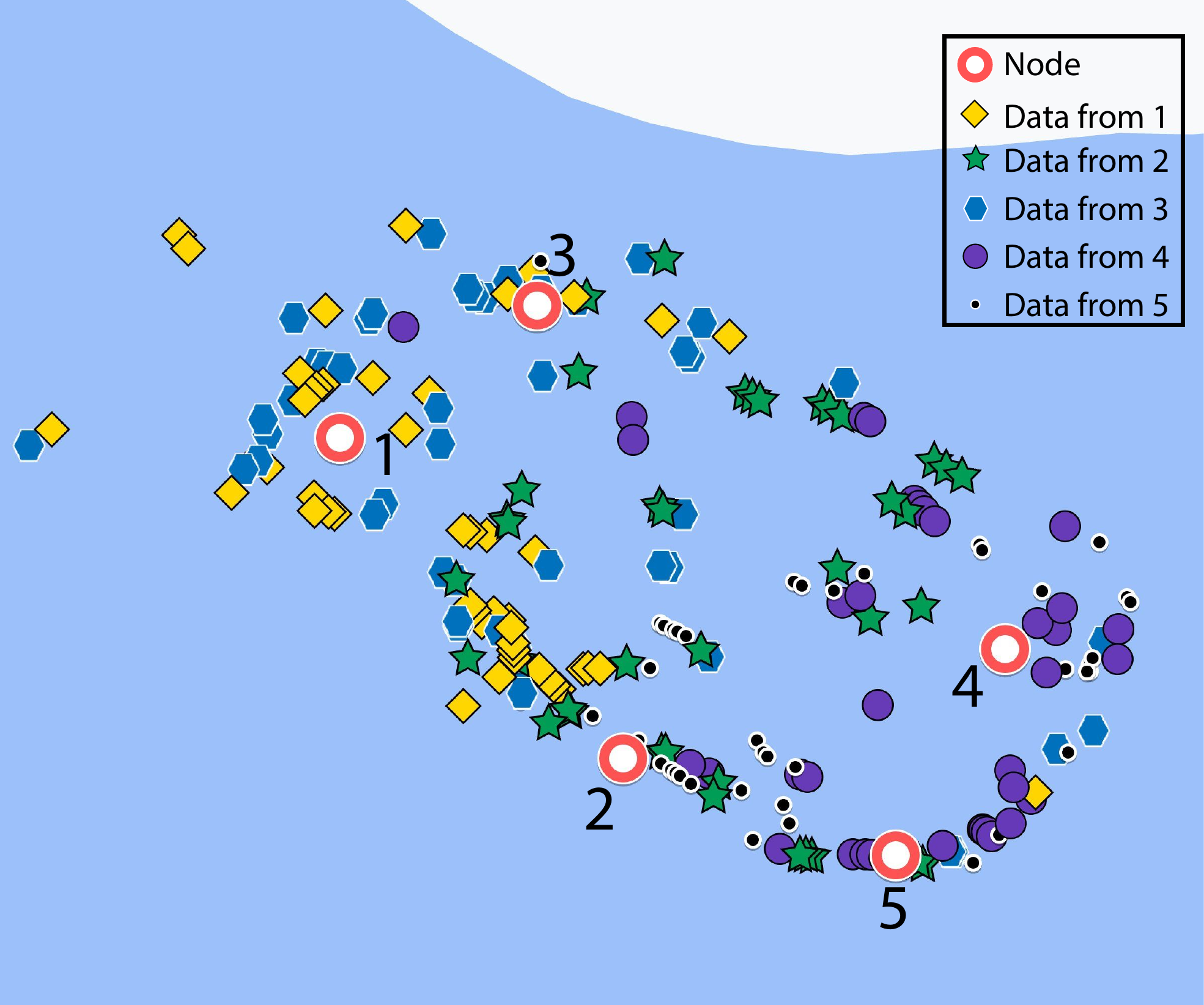}
        \caption{}
        \label{fig:top2_res}
    \end{subfigure}\\
    \begin{subfigure}{.5\textwidth}
        \centering
        \includegraphics[width=.95\linewidth]{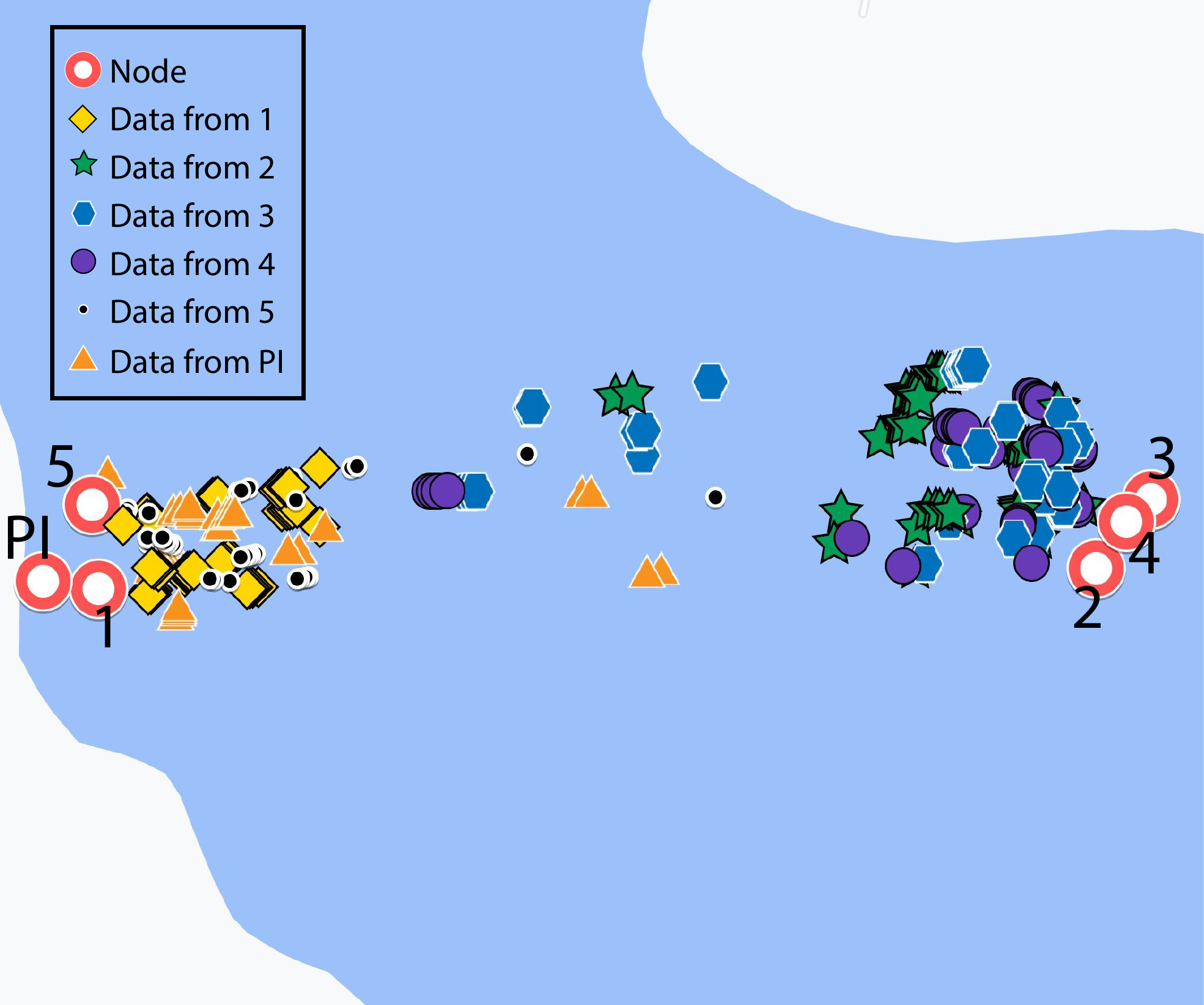}
        \caption{}
        \label{fig:top3_res}
    \end{subfigure}%
    \begin{subfigure}{.5\textwidth}
        \centering
        \includegraphics[width=.95\linewidth]{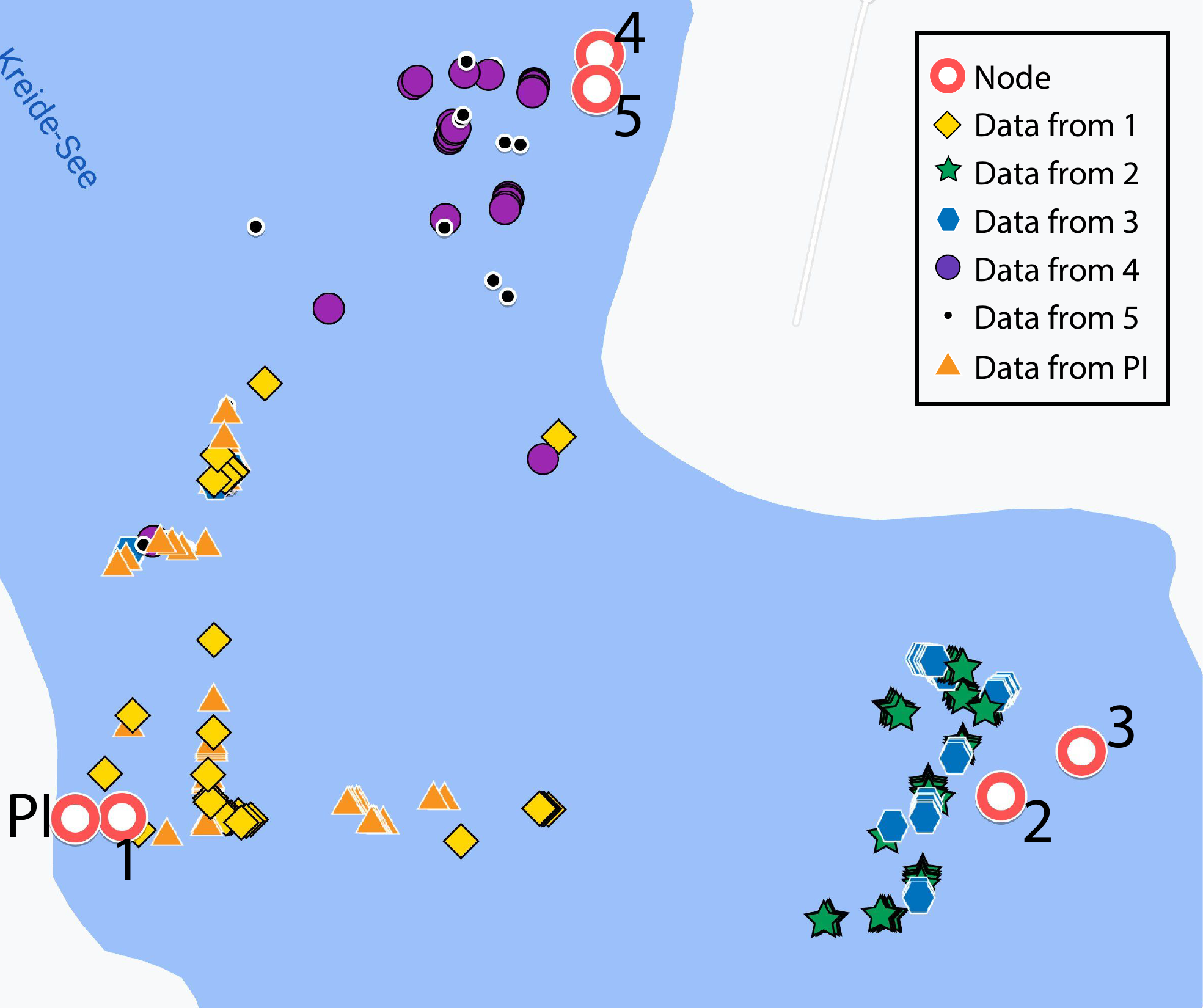}
        \caption{}
        \label{fig:top4_res}
    \end{subfigure}
    \caption{\gls{asv}'s positions for each received packet for all topologies.
    }
    \label{fig:topALL_res}
\end{figure}

Figures~\ref{fig:topALL_res} dig deeper into the relation between the \gls{asv} positions and the received packets, showing the actual \gls{asv} position for each received packet from each node in all the four analyzed topologies.
In the figures, the packets received by the \gls{asv} have a different marker for each transmitting node, i.e., yellow diamonds for the packets received from Node~1, green stars for those received from Node~2, blue hexagons for the packets received from Node~3, purple circles for the packets received from Node~4, black dots for the packets received from Node~5 and  orange triangles for the packets received from Node~6.
In topologies where the distance between nodes or between clusters of nodes is greater than the maximum transmission range, the cluster formation is well defined in proximity of the interested nodes, like in Topologies~1, 3 and 4.
Specifically, in Figure~\ref{fig:top1_res} we can observe the network behavior in Topology~1, where the \gls{asv} can receive the packets only when it is in the proximity of a single node, hence the derived figure shows a strong clustering of the packet receptions. 
Instead, in Topology~2 depicted in Figure~\ref{fig:top2_res}, the reception of packets by the sink node can happen even when the \gls{asv} was on the opposite side of the network, meaning that all the nodes are in range during most of the experiment, depending on the instantaneous channel conditions.
In Topology~3, presented in Figure~\ref{fig:top3_res}, the cluster formation is divided in two regions that correspond to the two main groups of nodes at the opposite sides of the lake. Some of the packets are received even in the middle of the lake since the \gls{asv} was still in range during their transmission.
Finally, Figure~\ref{fig:top4_res} shows the received packets in Topology~4. Also in this case the three clusters are well defined. Still, there are some packets received even during the travel of the \gls{asv} from one cluster to another, as in Topology~3, but this occurs less often than in the previous case, especially between the group 4-5 and group 2-3, since the \gls{asv} was traveling in a clockwise path between the different groups. Thus, group 2-3 is shielded by the natural headland until the \gls{asv} reached at least the middle of that path segment.

Finally, the JFI for each topology is reported in Table~\ref{tab:fairness}: the closer the index to one, the more fairly the protocol treats the nodes of the network. The results show that all nodes are treated fairly by the polling protocol, since the JFI ranges from 0.976 to 0.987 for all network topologies.

\begin{table}[!t]
\centering
\caption{Fairness Index for every topology}
\label{tab:fairness}
\begin{tabular}{@{}ccccc@{}}
\toprule
         & Topology 1 & Topology 2 & Topology 3 & Topology 4 \\ \midrule
JFI & 0.9755     & 0.9869     & 0.9874     & 0.9842     \\ \bottomrule
\end{tabular}
\end{table}
\section{Conclusions and Further Work} \label{sec:concl}

This paper presents a detailed procedure for enabling the usage of robotic systems, such as, but not limited to, \glspl{asv} or \glspl{rov} in industrial applications with modern web-based visualization tools and a robust communication infrastructure. Starting from the conclusions obtained in~\cite{deleacommunication} in the design and implementation phases, the overall communication  infrastructure system has been tested in simulation and deployed for lake trials. \remembertext{trl}{\fc{This has first of all validated the communication  infrastructure concept for \glspl{trl} 4 and 5 and enabled proceeding to the final stage for obtaining \gls{trl} 5-6 through a public demonstration in an industrial environment, a task that was accomplished with the live demo performed during the ITS World Congress in October 2021 in the Port of Hamburg~\cite{its2021}.}}

While the above water and underwater communications have been discussed in detail as the theme of this work, the presented results can and should be considered in the overall context of the \gls{robovaas} project. The experimental results presented were all collected using the service architecture successfully integrated for the four use-cases presented in Section~\ref{sec:use_case}. This includes the web user interface, or webUI, where each topology run was set up by a job request, then processed via the back-end applications supporting the \gls{robovaas} service and assigned to the SeaML as the \gls{asv} to perform the job. All the data was saved in the \gls{robovaas} database and is available for post processing. The SeaML \gls{asv}, developed and intended as the modular demonstration platform, was successfully deployed for the 10 days of the test campaign and performed all the required tasks.

The introduction of web services enabled rapid access to experimental data by multiple devices, regardless of their operating system or processor architecture, while still maintaining control over the access rights into the database and keeping the robotic systems secure from hacking. \remembertext{hacking}{\fc{The security is obtained, firstly, through logic enforcement done by the back-end services, which authenticate each request via a secret token generated at login and through exposing the front-end server port over secured web-sockets. With the exception of the Front Server, all other server-side components are accessible solely within local area network, where trusted clients are assumed. The clients within the local area network are assumed to be given access by the services provider, while the \gls{wpa2} encrypted wireless link that connects the robotic system being monitored only allows a limited number of devices to simultaneously access the system (in the current sea trials this number was limited to \SI{20}{}).}} The \gls{rmq} broker applied the access rights to the exchanges and queues dynamically enforced by the central server. This procedure shifted the focus to the roles of each user, because any mission required to have an operator and a robot assigned in order for the service to be started. This in turn enabled rapid development of all use-cases within the \gls{robovaas} project. The use-case definition has to be incorporated into the high-level control system of the robotic system, inside the central server's logic and inside the operator's standalone application. This process could be automated through message templates, similar to the message type primitives characteristic of \gls{ros}-based systems.

By using a non-SQL database such as MongoDB, the advantage in terms of flexibility of data insertion, migration or export was counterbalanced by the lower speed of the queries, which needed to parse several thousand JSON objects for each request for historic data. \fc{ This flexibility allowed loosely defining the database models and relying more on data parsing. This is particularly useful when complex structures, with multiple parameters in exchanged messages are nested within one another. This has the drawback that, by increasing the complexity of the data that needs to be parsed, the performance of the queries decreases significantly: other solutions based, for instance, on SQL databases, where models are defined a priori, can overcome this issue at the price of a lower flexibility.}

On the quality of service performed with unmanned vehicles, the data-muling use-case successfully combined underwater and above water communication technologies deployed on an \gls{asv}. In comparison to deployment on \glspl{uuv}, the surface vehicles can cover large areas in short amounts of time like \glspl{auv}, while still maintaining reasonably high runtime autonomy. 

Having benchmarked the performance of the communication system and validated the overall system through sea trials, the proposed solution reached \gls{trl} 5-6. A future  goal is to validate the system in an industrial environment and therefore reach \gls{trl} 6-7. In parallel, the high-level control system of the \gls{asv} will be restructured to be \gls{ros}-compliant and augment different control strategies to enhance the autonomy of the robot.
Future work will also focus on improving the packet delivery ratio of the acoustic modems in a mobile network where the propellers noise causes a significant performance drop. 
To this aim, different modulation and synchronization techniques will be investigated, to obtain a higher resilience against propellers noise.
Lastly, additional relevant use-cases, such as marine litter cleaning or infrastructure inspection, will be investigated to provide robotic assistance solutions with the help of the modular platform developed and with the above water and underwater communication system infrastructure.

\section*{Acknowledgements}

This work has been partially supported by the Italian Ministry of Education, Universities and Research (MIUR), the German Federal Ministry of Economy (BMWi) and ERA-NET Cofund MarTERA (contract~728053).

The authors would like to thank Christian Busse (Hamburg University of Technology) for his support to build the underwater nodes.

\bibliographystyle{IEEEtran}
\bibliography{library/refen,library/smp_ref,library/cml_ref}

\newpage

\end{document}